\documentclass[sigconf]{acmart}
\pdfoutput=1
\usepackage{booktabs} 
\usepackage{multirow}

\usepackage{flushend}
\usepackage{hyperref}

\usepackage{dsfont,amsfonts,amssymb,amsmath,color}
\usepackage{latexsym}
\usepackage{amsmath}
\usepackage{amsfonts}
\usepackage{graphicx}

\usepackage[ruled,vlined,linesnumbered,lined, commentsnumbered]{algorithm2e}

\usepackage{subfigure}
\usepackage{caption}
\theoremstyle{definition}
\newtheorem{definition}{Definition}[section]

\newcommand{\yq}[1]{\textcolor{black}{#1}}
\newcommand{\hr}[1]{\textcolor{black}{#1}}
\newcommand{\ph}[1]{\textcolor{black}{#1}}

\makeatletter
\def\subsubsection{\@startsection{subsubsection}{3}%
  \z@{.5\linespacing\@plus.7\linespacing}{.1\linespacing}%
  {\normalfont\itshape}}
\makeatother

\usepackage[ruled,vlined]{algorithm2e}
\usepackage{graphicx}
\usepackage{subfigure}

\copyrightyear{2021} 
\acmYear{2021} 
\setcopyright{acmcopyright}\acmConference[CIKM '21]{Proceedings of the 30th ACM International Conference on Information and Knowledge Management}{November 1--5, 2021}{Virtual Event, QLD, Australia}
\acmBooktitle{Proceedings of the 30th ACM International Conference on Information and Knowledge Management (CIKM '21), November 1--5, 2021, Virtual Event, QLD, Australia}
\acmPrice{15.00}
\acmDOI{10.1145/3459637.3482252}
\acmISBN{978-1-4503-8446-9/21/11}

\begin{document}
\title{Differentially Private Federated Knowledge Graphs Embedding}

\author{Hao Peng$^{1,4}$, Haoran Li$^{2,5}$, Yangqiu Song$^{2,5}$, Vincent Zheng$^{3}$, Jianxin Li$^{1,6}$}
\affiliation{
\institution{
$^1$Beijing Advanced Innovation Center for Big Data and Brain Computing, Beihang University, Beijing 100191, China;\\
$^2$Department of Computer Science and Engineering, HKUST, Hongkong, China;
$^3$AI Group, Webank Co., Ltd;\\
$^4$School of Cyber Science and Technology, Beihang University, Beijing 100191, China;\\
$^5$Peng Cheng Laboratory, Shenzhen 518066, China;
$^6$SKLSDE, Beihang University, Beijing 100191, China;\\
}
}
\email{{penghao, lijx}@act.buaa.edu.cn, hlibt@connect.ust.hk, yqsong@cse.ust.hk, vincentz@webank.com.}
\thanks{Hao Peng and Haoran Li contribute equally.}

\renewcommand{\shortauthors}{Peng and Li, et al.}

\begin{abstract}
Knowledge graph embedding plays an important role in knowledge representation, reasoning, and data mining applications.
However, for multiple cross-domain knowledge graphs, state-of-the-art embedding models cannot make full use of the data from different knowledge domains while preserving the privacy of exchanged data. 
In addition, the centralized embedding model may not scale to the extensive real-world knowledge graphs.
Therefore, we propose a novel decentralized scalable learning framework, \emph{Federated Knowledge Graphs Embedding} (FKGE), where embeddings from different knowledge graphs can be learnt in an asynchronous and peer-to-peer manner while being privacy-preserving. 
FKGE exploits adversarial generation between pairs of knowledge graphs to translate identical entities and relations of different domains into near embedding spaces. 
In order to protect the privacy of the training data, FKGE further implements a privacy-preserving neural network structure to guarantee no raw data leakage.
We conduct extensive experiments to evaluate FKGE on 11 knowledge graphs, demonstrating a significant and consistent improvement in model quality with at most 17.85\% and 7.90\% increases in performance on triple classification and link prediction tasks.
\end{abstract}

%
%



\maketitle

\renewcommand{\thefootnote}{\fnsymbol{footnote}}

\section{Introduction}\label{sec:intro}
Knowledge graphs (KGs) have been built to benefit many applications, e.g., semantic search, question answering, recommendation systems, etc. ~\cite{ji2020survey,EhrlingerW16,yang2021kgsynnet,liu2021kg}.
There have been several big and general-purpose KGs such as Freebase ~\cite{freebase} (or later Wikidata ~\cite{vrandevcic2014wikidata}) and Yago~\cite{YAGO}, and numerous domain-specific KGs of various sizes such as GeoNames ~\cite{ahlers2013assessment} in geography and Lexvo ~\cite{de2015lexvo} in linguistics.
To our best knowledge, most companies build their own commercial KGs, which usually require laborious human annotation and high computational cost\footnote{See \url{https://www.maana.io/} for a general pipeline of components.}.
However, there are multiple reasons that companies would not want to share their KGs.
First, each company has its private part of data, which cannot be disclosed to others.
Second, even if privacy is not a concern, they would not expose their knowledge to other companies except they can also benefit from others.
Third, integrating knowledge itself is not trivial or easy. 
On the other hand, in many cases, companies indeed have the motivation to exchange knowledge to improve their own data quality and service. 
For example, a drug discovery company may benefit from a patient social network and a gene bank, and so as the other two interested in a drug network and a disease network owned by the drug discovery company.
Sometimes, the data cannot even be bought due to the privacy concern.
For example, under EU's GDPR\footnote{\url{https://gdpr-info.eu}}
, companies cannot use or share a user's data without his/her consent.
Therefore, a more loosely coupled and principled way to share their KGs that can benefit multiple parties should be considered.




Traditionally, federated database systems ~\cite{ShethL90} were proposed to support unified query language over heterogeneous databases without doing actual data integration.
Such systems do not help improve individual KG's quality or service with private data preserved.
Recently, machine learning, in particular KG representation learning or KG embedding, has been shown to be powerful for knowledge representation, reasoning, and many downstream applications \cite{Nickel0TG16,wang2017knowledge,ji2020survey,mao2021event}.
When representing entities and their relations in a vector space, different KGs may share information if their embedding spaces are (partially) aligned ~\cite{ChenTYZ17,zhu2017iterative,sun2020knowledge,xu2020coordinated}.
\yq{However, revealing vector representations to other parties can also leak private information~\cite{carlini2020extracting}.
Currently,} for multiple cross-domain KGs, especially for the ones shared by different companies, state-of-the-art KG embedding models cannot make full use of the data from different domains while preserving the privacy.
More recently, {\it federated machine learning} ~\cite{YangLCT19} has been widely considered when heterogeneous devices or multiple parties contribute non-IID data where privacy of different sources should be preserved.
However, the natures of distributed federated learning ~\cite{konevcny2016federated,konevcny2016federated2} (which requires centralized averaging platform) and vertical or horizontal federated learning (which do not consider graph structures) make them non-trivial to be adapted to KG embeddings.
There are some existing federated learning mechanisms for graphs ~\cite{Lalitha2019_p2p} but they treat each node to be a computing cell and cannot collaboratively train multiple KGs together.

\begin{figure}[t]
\center
\includegraphics[width=0.46\textwidth]{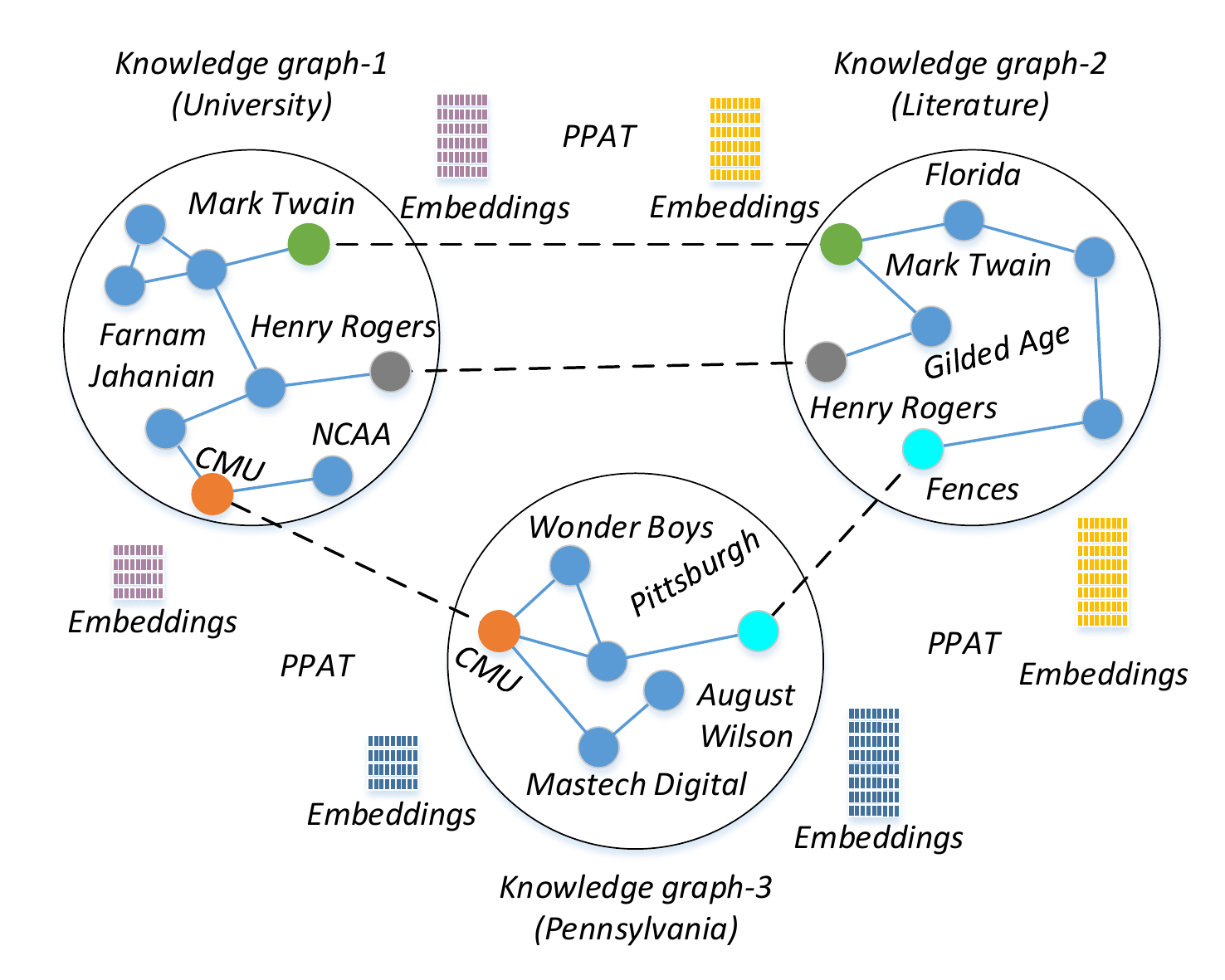}\vspace{-0.15in}
\caption{Overview of FKGE framework. Different Knowledge graphs may use different embedding models.}\label{fig:KGs}\vspace{-0.15in}
\end{figure}

To improve the quality of individual KGs for multiple cross-domain KGs, we propose a new framework called \underline{F}ederated \underline{K}nowle-dge \underline{G}raphs \underline{E}mbedding (FKGE), where embeddings from different KGs can be learnt jointly in an asynchronous and pairwise manner while being privacy-preserving.
In FKGE, we design a \ph{privacy-preserving adversarial translation (PPAT) network} to \yq{mutually enhance} embeddings of \yq{each pair of KGs based on} aligned entities and relations.
\ph{PPAT network's} \yq{mechanism guarantees the differential privacy for the paired embeddings of aligned entities and relations not leaked to each other.}
For example, in Fig. ~\ref{fig:KGs}, \emph{Mark Twain} and \emph{Henry Rogers} are two of aligned entities that belong to both {\it University} and {\it Literature} KGs, they have different representations in two embedding spaces of two KGs.
\yq{We use \ph{PPAT network} to separate their embeddings from both sides \hr{while} we are still able to \hr{exploit their embeddings to} improve the embedding quality for both  {\it University} and {\it Literature} KGs.}

In summary, we highlight the following characteristics of our FKGE
\footnote{Code is available at \url{https://github.com/HKUST-KnowComp/FKGE}} 
framework.
1) FKGE framework is asynchronous and decentralized.
Different from centralized client-based models, FKGE pairs up KGs from different domains with an adversarial network.
2) FKGE is scalable and compatible with many base embedding models. 
The asynchronous and decentralized setting leads to parallel computation between pairs of collaborators. 
Moreover, FKGE can serve as a meta-algorithm for existing KG embedding methods through a handshake protocol.
3) FKGE is privacy-preserving and guarantees no raw data leakage.
FKGE's design requires no raw data transmission between collaborators, and transmitted generated embeddings are differentially private.
\section{Related Work}\label{sec:relat}
As our work is closely related to federated learning and knowledge graph embedding, the related works are two-fold.

\subsection{Federated Learning}
Federated learning allows multiple data owners to collaborate on building models without compromising on data privacy.
Google first proposed a federated learning framework between the cloud server and edge devices to train on edges and update global model on the cloud~\cite{konevcny2016federated,konevcny2016federated2,mcmahan2017communication}.
Yang et al. ~\cite{YangLCT19} further extended the definition from edge devices to general data owners, so that the collaborations between separate databases were taken into scenarios of federated learning.
For graph structured data, Lalitha et al. ~\cite{Lalitha2019_p2p} considered using federated learning in a peer to peer manner.
FedE ~\cite{chen2020fede} exploited federated learning over a KG through centralized aggregation for the link prediction task.
However, both of them \yq{handled one single graph by either treating each node to be a computing cell or distributing triplets in a KG into different servers and performed the Federated Averaging algorithm introduced by Google~\cite{mcmahan2017communication}, and thus cannot collaboratively train multiple KGs together.}

In terms of privacy preserving mechanisms, most works used homomorphic encryption (HE) ~\cite{HE_paillier}, secure multi-party computation (MPC) ~\cite{Mohassel17_MPC}, and differential privacy (DP) ~\cite{Dwork08,Robin2017,grammenos2020federated} to improve security.
There are also serval extensions which combine or improve the above frameworks.
For example, Lyu et al.~\cite{lyu2020towards} used DP, HE, and Blockchain technologies to build a decentralized fair and privacy-preserving deep learning framework.
Xu et al.~\cite{xu2019verifynet} designed a verifiable federated learning framework based on the homomorphic hash function and pseudorandom.
Our proposed FKGE framework is targeted for multiple KGs with millions of entities and HE is inefficient for our task.
DP addresses individuals' privacy concerns while aggregating different databases so that specific users' information can not be inferred through federation.
Therefore, DP algorithm is implemented for our FKGE.




\subsection{Knowledge Graph Embedding}
KG embedding plays an important role in knowledge base inferences and downstream applications~\cite{Nickel0TG16,wang2017knowledge,bianchi2020knowledge,li2020modeling}.
Popular models are mostly based on a translational model, where a head is translated to a tail through a relational embedding~\cite{bordes2013translating}, or a bilinear model, where a bilinear matrix is used to combine the head and tail to form a loss function~\cite{NickelTK11}.
Recently, there have been many extensions based on these two models~\cite{wang2014knowledge,lin2015learning,ji2015knowledge,xiao2016one}.
More recently, these models are extended to the complex space instead of using the Euclidean space~\cite{trouillon2016complex,sun2019rotate}, which can model many important properties of relations such as symmetry, inversion, and composition.

When there are multiple KGs, embedding based entity alignment can be performed. 
Such entity alignment task usually assumes that different KGs are partially aligned and tries to predict more aligned entities.
For example, the entity alignment can be based on cross-lingual KGs~\cite{ChenTYZ17,ZhuZ0TG19,WuLF0Y019,Pei0Y020}, KGs with multi-view entity-related information~\cite{ChenTCSZ18,zhang2019multi,CaoLLLLC19}, and KGs in similar domains with significant entity overlaps~\cite{zhu2017iterative,sun2018bootstrapping,TrisedyaQZ19,YangLZWX20}.
After joint learning, embeddings for entity alignment are usually aligned in a unified space so the vectors can be used to find nearest entities in other KGs.
Note that our FKGE framework is different from above KG alignment problems.
Instead of predicting the potential aligned entities using given ones, we aim to improve individual KG embeddings based on provided aligned entities.
In FKGE, after joint training, each KG still does not know other KG's embedding space, but embeddings in each KG are all improved for better downstream tasks such as node classification or link prediction.
This is guaranteed by the differential privacy mechanism that we introduced in our \ph{PPAT network}: when training each pair of embedding sets for the aligned entities, they cannot leak a single embedding since inclusion and exclusion of a particular embedding will not affect the out-come distribution very much.
This also allows us to use different base KG embedding models for different KGs.

\begin{figure*}[t]
\centering
\vspace{-0.25in}
\includegraphics[width=0.95\textwidth]{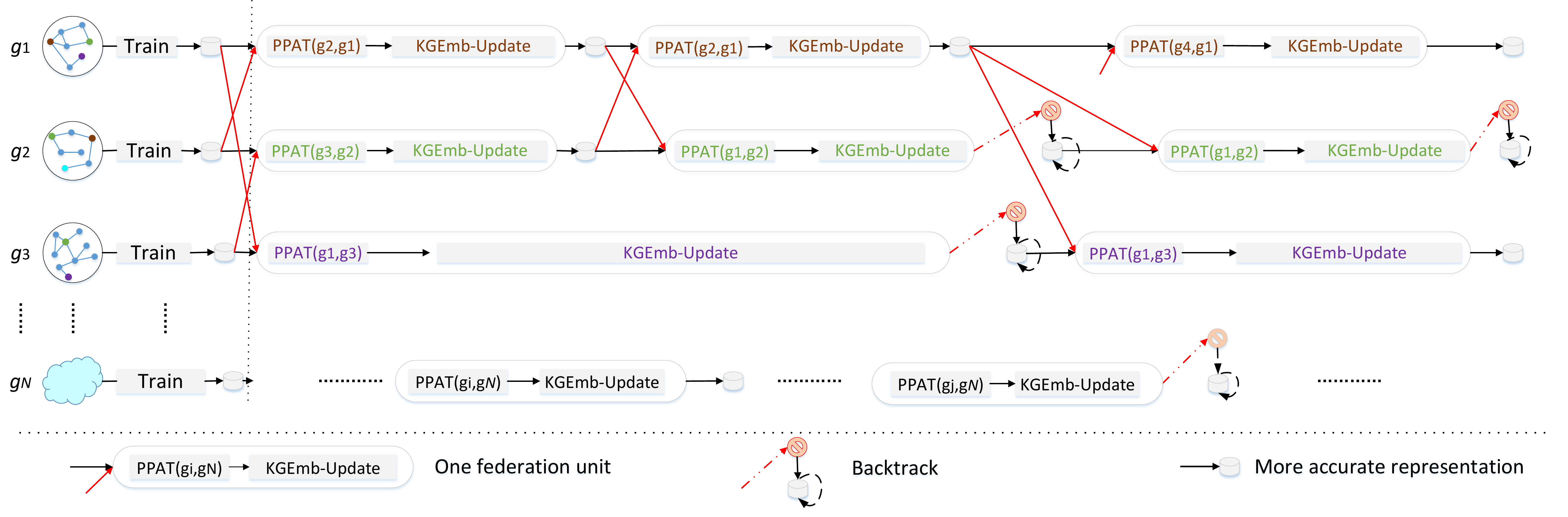}\vspace{-0.15in}
\caption{An example of whole training procedure.
The order of training is not fixed and depends on individual KG owner's computation power and willingness to cooperate.
``Train'' indicates training KGE model locally. 
``\ph{PPAT}$(g_i,g_j)$'' denotes the \ph{PPAT} embeddings between $g_i$ and $g_j$ where the generator locates in client $g_i$ and the discriminator lies in host $g_j$. 
``KGEmb-Update'' updates aligned embeddings with generated output of \ph{PPAT network} and retrains all the embeddings as ``Train''.
``Backtrack'' happens if evaluation result after ``KGEmb-Update'' is not improved, then newly trained embeddings are abandoned and previous embeddings are kept.
Otherwise the result is backtracked.
}\label{frame}
\vspace{-0.15in}
\end{figure*}

\section{Federated Knowledge Graphs Representation Learning}\label{sec:model}
In this section, we present detailed descriptions of FKGE.
We first give the problem definition and sketch intuitive solution in Section~\ref{sec:Problem Formulation}.
Then we introduce more details of our privacy-preserving model in Section \ref{sec:PPAT}.
Finally, we give a comprehensive explanation of our federated training mechanism in Section \ref{sec:fed_train}.

\subsection{Problem Formulation and Proposed Solution}\label{sec:Problem Formulation}
We define the set of knowledge graphs (KGs) from separate owners as $\mathcal{KG} = \{g_1, g_2, g_3, ..., g_N\}$, 
where $N$ is the total number of KGs.
Every element in $\mathcal{KG}$ locates in different databases and cannot access other KGs' databases.
Let $g_k = \{\mathcal{E}_k, \mathcal{R}_k, \mathcal{T}_k\} (1 \leq k \leq N)$ denote the $k$-th KG in $\mathcal{KG}$.
Each triple $(h,r,t) \in \mathcal{T}_k$ (the set of triples of $g_k$) is composed of a head entity $h \in \mathcal{E}_k$ (the set of entities of $g_k$), a tail entity $t \in \mathcal{E}_k$ and a relation $r \in \mathcal{R}_k$ (the set of relations of $g_k$). 
For any pair of KGs $(g_i, g_j)$ in $\mathcal{KG}$, we assume that both aligned entities $\mathcal{E}_i \cap \mathcal{E}_j$ and relations $\mathcal{R}_i \cap \mathcal{R}_j$ are given which can be done by a secure hash\footnote{\url{https://csrc.nist.gov/publications/detail/fips/180/4/final}}.
Our goal is to exploit aligned entities and relations to further improve all embeddings of any individual KG.
To aid discussion, Tab. ~\ref{tab:symbol} depicts the notations used throughout the paper.
Each KG owner $g_i$ trains its own embeddings of entities $\mathcal{E}_i$ and relations $\mathcal{R}_i$ locally. 
Based on the trained embeddings, FKGE aggregates the embeddings of both aligned entities and relations from paired KGs, and then updates embeddings in a federated manner.
For aligned entities and relations from any pair of KGs, e.g., $(g_i, g_j)$, FKGE includes a secure pipeline that can refine the embeddings of $\mathcal{E}_i \cap \mathcal{E}_j$ and $\mathcal{R}_i \cap \mathcal{R}_j$ and further improve embeddings of \hr{$\mathcal{E}_i \cup \mathcal{R}_i$ and $\mathcal{E}_j \cup \mathcal{R}_j$} individually.
Moreover, FKGE proposes a federated training mechanism to improve all the parties jointly via broadcasting.
If $g_i$ or $g_j$ gains improvement from the refined embeddings, it will broadcast signals to other KGs to further improve overall results.
\begin{table}[t]
\footnotesize
    \centering
    \caption{Glossary of Notations.}\label{tab:symbol}
    	\vspace{-1em}
    \setlength{\tabcolsep}{3mm}
    \begin{tabular}{r|l}
    \toprule
   \textbf{Symbol} & \textbf{Definition} \\
    \midrule
    $\mathcal{KG}$, $N$, $g_k$ & Knowledge graph set, its size, $k$-th knowledge graph\\
    $\mathcal{E}_k$,  $\mathcal{R}_k$, $\mathcal{T}_k$ & The set of entities, relations, triples of $g_k$ \\
    $X$, $Y$ & Embedding of aligned entities and relations of client, host \\
    $n$ & Total number of training samples\\
    $d$ & Embedding dimension of entity and relation \\
    $\mathcal{N}(X)$ & Raw embeddings of neighbor entities and joining relations of $X$\\ 
    \midrule
    $S$, $T_i$ & Student discriminator, $i$-th teacher discriminator\\
    $\theta_S$ , $\theta_T^{i}$ & Parameters of student, $i$-th teacher discriminator\\
    $G$, $\theta_G$ & Generator, parameters of generator that includes the mapping\\ 
    &  matrix $W$\\

    $|T|$ & Total number of teacher discriminators\\
    \midrule
    $\lambda$ & Noise (scale) of Laplace random variable\\
    $n_0$, $n_1$ & Total vote number for 0,1 of teacher discriminators\\
    $l$ & A chosen positive integer for moment \\

    \bottomrule
    \end{tabular}
\end{table}
Otherwise, it will backtrack to original embeddings before federation.
As an example shown in Fig. ~\ref{frame}, in the beginning, $g_1$, $g_2$, $g_3$ train their embedding locally. 
During first federation, they form 3 pairs of KGs: $(g_1,g_3)$, $(g_2,g_1)$ and $(g_3,g_2)$.
After first federation, $g_1$ and $g_2$ gain improvement for overall embeddings. 
$g_3$'s training takes longer time and fails to improve its embedding, therefore $g_3$ backtracks to initial embedding.
During second federation, $g_1$ and $g_2$ pair up as $(g_2,g_1)$ and $(g_1,g_2)$ and only $g_1$ gains improvements.
$g_2$ backtracks to previous embedding. 
Since $g_3$ is still on the training process, it will not join second federation and will go to sleep state if no available KG exists.
For third federation, $g_1$ finishes its training and broadcasts $g_3$ to wake up. 
Then they form $(g_1,g_3)$, $(g_1,g_2)$ and $(g_4,g_1)$ pairs for federation.
The whole training procedure continues until no more improvement for all the KGs.



\subsection{Privacy-Preserving Adversarial Model}\label{sec:PPAT}
Given two KGs $(g_i, g_j)$ with aligned entities $\mathcal{E}_i \cap \mathcal{E}_j$ and relations $\mathcal{R}_i \cap \mathcal{R}_j$, FKGE exploits a GAN  ~\cite{GoodfellowPMXWOCB14} structure to \hr{unify the embeddings of aligned entities and relations, where we borrow the idea of MUSE ~\cite{gan_translation} which used a translational mapping matrix $W$ in the generator to align two manifolds in the GAN's intermediate vector space}. 
The generator and the discriminator can locate in different sides for a pair of KGs.
More specifically, we may put the generator in $g_i$ and the discriminator in $g_j$.
The generator translates aligned entities' embeddings from $g_i$ to $g_j$ and the discriminator distinguishes between synthesized embeddings of the generator and ground truth embeddings in $g_j$.
After GAN training, the synthesized embeddings are able to learn features from both KGs and therefore can replace \hr{original embeddings of $\mathcal{E}_i \cap \mathcal{E}_j$ and $\mathcal{R}_i \cap \mathcal{R}_j$} in $g_i$ and $g_j$ as refined and unified embeddings.
It is sufficient for GAN training that only the generated outputs and gradients are transmitted between $g_i$ and $g_j$ without revealing raw data.
However, even for generated embeddings, there are still privacy concerns for reconstruction attacks.
It is possible that neural models may memorize inputs and reconstruct inputs from corresponding outputs~\cite{carlini2020extracting}.
To further address the privacy issue, we introduce differential privacy to privatize generated embeddings.
Differential privacy provides a strong guarantee for protecting any single embedding in the generator outputs since inclusion and exclusion of a particular embedding will not affect the outcome distribution very much.
Definitions~\ref{def:neighbor data} and ~\ref{def:dp} give the formal definition of differential privacy.

\begin{definition}[Neighboring Dataset]
\label{def:neighbor data}
Two datasets $D, D'$ are neighboring if 
\begin{equation}
\small
\exists x \in D \; s.t.\; D-\{x\}=D'.
\end{equation}
\end{definition}

\begin{definition}[Differential Privacy]
\label{def:dp}
A randomized \textit{algorithm mechanism} $M$ with domain $D$ and range $R$ satisfies $(\epsilon,\delta)$-\textit{differential privacy} if for any two neighboring datasets $D, D'$ and for any subsets of output $O \subseteq R$:
\begin{equation}\label{eq:dp-bound}
\small
Pr[M(D) \in O] \leq e^{\epsilon}Pr[M(D')\in O]+\delta.
\end{equation}
\end{definition}
\noindent
The $\epsilon$ corresponds to privacy budget. 
Smaller $\epsilon$ asserts better privacy protection and lower model utility since algorithm outputs of neighboring datasets are similar. 
$\delta$ is the probability of information accidentally being leaked. 
Relying on above definitions, PATE-GAN ~\cite{yoon2018pategan} proposed a revised GAN structure to generate differentially private generator outputs by applying PATE mechanism \cite{nicolas_pate} with teacher and student discriminators.
Previously, PATE-GAN ~\cite{yoon2018pategan} was tested on simple image classification tasks, where all images were in one domain.
Putting the translational mapping matrix $W$ in the generator, we are able to apply PATE-GAN structure for the complicated cross-domain KGs.

Based on above intuitions, we implement our \ph{privacy-preserving adversarial translation (PPAT) network}.

\subsubsection{Model Architecture and Loss Formulation}
\begin{figure}[t]
\centering
\includegraphics[width=0.5\textwidth]{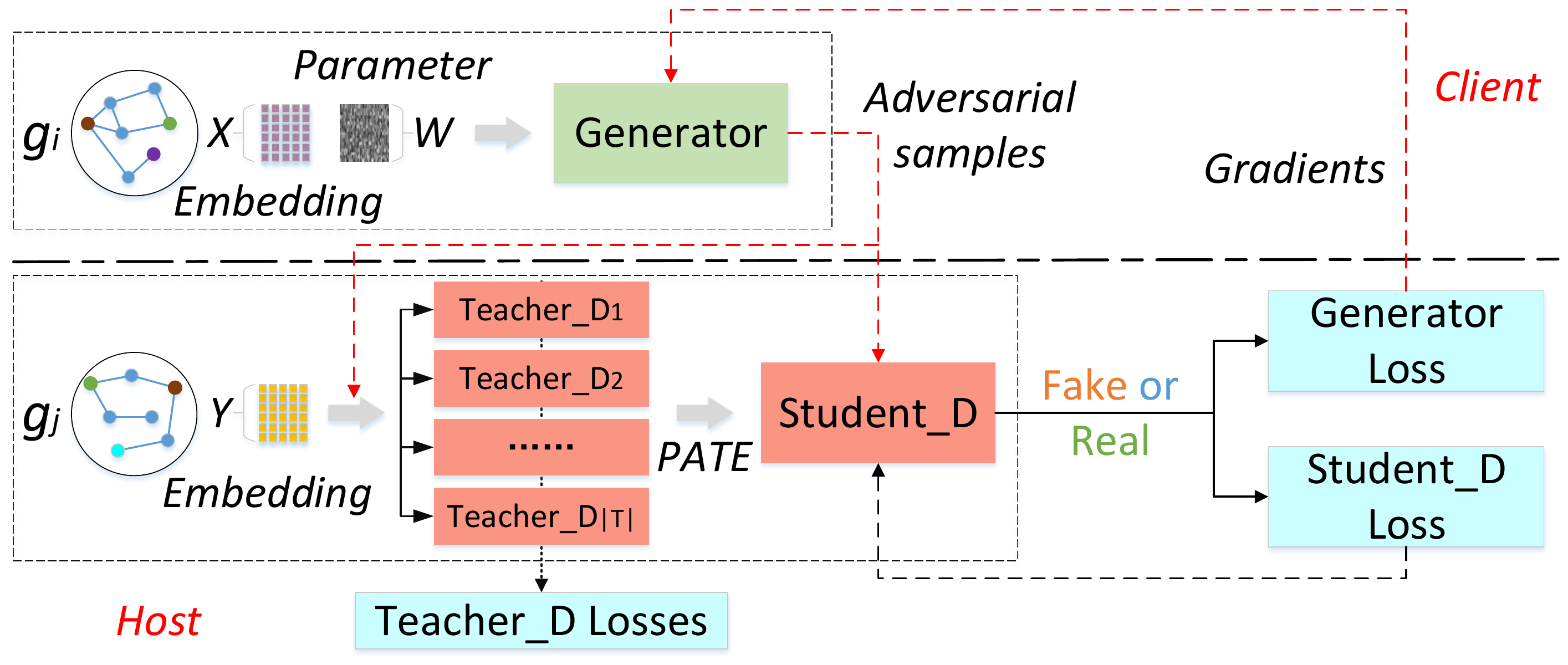}\vspace*{-2.5mm}
\caption{The architecture of \ph{PPAT} network. The host and client are separated: only adversarial samples and gradients from the host are exchanged, all raw data and embeddings are private. Here ``D'' in host is short for discriminator.}\label{GAN}
\vspace{-0.15in}
\end{figure}

For any pair of $(g_i, g_j)$, the architecture of our \ph{PPAT network} is illustrated in Fig.~\ref{GAN}.
\ph{PPAT network} exploits GAN structure to generate differentially private synthetic embedding with high utility.
It replaces the original GAN discriminator with multiple teacher discriminators and one student discriminator to achieve differential privacy of generated embeddings.
The generator $G$ with parameters $\theta_G$ \hr{(which also is the translational mapping matrix $W$ in Fig. \ref{GAN})} locates in $g_i$'s database while the student discriminator $S$ with parameters $\theta_S$ and multiple teacher discriminators $T=\{T_1,T_2,...,T_{|T|}\}$ with parameters $\theta_T^1,\theta_T^2,...,\theta_T^{|T|}$ lie in $g_j$'s database. 
We denote $g_j$ as host and $g_i$ as client, since the host is responsible for the generator's and all discriminators' loss back-propagation calculations, while the client only transmits its generated embeddings and receives gradients to update its generator parameters.
We use $X=\{x_1,x_2,...,x_n\}$ to denote embeddings for $\mathcal{E}_i \cap \mathcal{E}_j$ and $\mathcal{R}_i \cap \mathcal{R}_j$ in $g_i$, and $Y=\{y_1,y_2,...,y_n\}$ to denote embeddings for $\mathcal{E}_i \cap \mathcal{E}_j$ and $\mathcal{R}_i \cap \mathcal{R}_j$ in $g_j$.

The objective of the generator $G$ is to generate adversarial samples by making $G(X)$ and $Y$ similar so that the student discriminator $S$ cannot distinguish them.
Eq. (\ref{eq:G loss}) presents the objective function of the generator loss:
\begin{equation}
\label{eq:G loss}
\small
L_G(\theta_G ; S) = \frac{1}{n} \sum_{m=1}^{n} \log(1-S(G(x_m);\theta_G)),
\end{equation}
where \hr{$G(X) = W X$}, and $S$ is the student discriminator parameterized by $\theta_S$, which takes embeddings of both $G(X)$ and $Y$ as an input
\yq{under the CSLS metric used by MUSE ~\cite{gan_translation}.}

The learning objective of teacher discriminators is the same as the \hr{original} discriminator that distinguishes between fake samples $G(X)$ and real samples $Y$. 
The only difference is that teacher discriminators are trained on disjointly partitioned data. Teachers' losses are formulated as Eq. (\ref{eq:T loss}):
\begin{equation}
\label{eq:T loss}
\small
L_T^{i}(\theta_T^{i} ; G) = - [ \sum_{m=1}^{n}\log(1-T_i(G(x_m) ; \theta_T^i))
+\sum_{y_k \in D_i}\log (T_i(y_k ; \theta_T^i) )],
\end{equation}
where $D_i$ is partitioned subset consisted of the dataset $X$ and $Y$ for $T_i$ such that $|D_i| = \frac{n}{|T|}$ and there is no overlap between different subsets. 

The learning objective of the student discriminator $S$ is to classify generated samples given aggregated noisy labels.
More specifically, teacher discriminators' predictions together with randomly injected Laplace noises will determine the labels for the student discriminator $S$.
Eqs. (\ref{eqn:Noisy Aggregation}) and (\ref{eqn: n_j}) illustrate the PATE mechanism:
\begin{equation}
\small
\label{eqn:Noisy Aggregation}
PATE_\lambda(x) = \mathop{\arg\max}_{j \in \{0,1\} } (n_j(x)+V_j),
\end{equation}
where $V_0$, $V_1$ are i.i.d. $Lap(\frac{1}{\lambda})$ random variables that introduce noises to teachers' votes. 
$n_j(x)$ denotes the number of teachers that predict class $j$ for input $x$:
\begin{equation}
\label{eqn: n_j}
\small
n_j(x) = |\{T_i:T_i(x) = j\}| \text{\quad for } j = 0, 1.
\end{equation}
Then the student discriminator exploits generated samples with voted noisy labels (shown in Eq. (\ref{eqn:Noisy Aggregation})) to train itself in the host database. 
The student loss function is formulated as:
\begin{equation}
\label{eq: S loss}
\small
L_S(\theta_S ; T,G) = \frac{1}{n} \sum_{i=1}^{n} [\gamma_i \log S(G(x_i); \theta_S)+(1-\gamma_i) \log(1-S(G(x_i); \theta_S))],
\end{equation}
where $\gamma_i = PATE_\lambda(x_i)$, the noisily aggregated label voted by teacher discriminators. 

Besides the generated embedding $G(X)$, after the training process of \ph{PPAT network} is stable, the client generates embeddings for the adjacent entities of the aligned entities in $g_i$ and the joining relations as types of virtual entities and relations that are added into the host.
We use $\mathcal{N}(X)$ to denote the raw embeddings of adjacent entities and joining relations, and the generated embeddings of virtual entities and relations can be denoted as $G(\mathcal{N}(X))$.
These virtual entities and relations are only used for KGE, and will be removed before responding to other hosts.

\subsubsection{Privacy Analysis and Parameter Estimation}
For the \ph{PPAT network}, our dataset $X$ to feed the generator is embedding of $\mathcal{E}_i \cap \mathcal{E}_j$ and $\mathcal{R}_i \cap \mathcal{R}_j$ in $g_i$, the neighboring dataset is defined by excluding a particular embedding from $X$.
We use $X'$ to denote the neighboring dataset such that there is an embedding $x_i \in X$ and $X-\{ x_i\}=X'$.
The algorithm to perform parameter estimation is shown in Algo.~\ref{DE_algo}.
The raw embeddings of aligned entities and relations in $X$ are fed to the generator to generate adversarial samples $G(X)$ which are later on transmitted to all teacher discriminators $T$ of the host.
By adding Laplace noise to voted results of teacher discriminators, differential privacy requirement is satisfied \cite{nicolas_pate}.
The student discriminator $S$ is then trained by the synthesized embeddings with aggregated labels that contains 0 or 1 voted by teacher discriminators.
The Post-Processing Theorem \cite{post_processing_thm} which states the composition of a data-independent mapping $f$ with an $(\epsilon, \delta)$ - differentially private algorithm $M$ is also $(\epsilon, \delta)$ -  \textit{differentially private}.
By Post-Processing Theorem, the student discriminator $S$ is also differentially private since it is trained by differentially private labels.
\hr{Moreover, the generator $G$ is differentially private since $G$ is trained by student discriminator $S$.}
\hr{Hence}, the transmitted embeddings are synthesized and differentially private since they are outputs of the generator $G$.

During the training process, the host calculates the generator's and all discriminators' loss functions locally: gradients of student discriminator loss and teacher discriminator losses are used to update discriminators' parameters locally, while gradients of generator loss are sent back to the generator to update its parameters.
Thus, neither of $g_i$ and $g_j$ is able to access the embeddings or raw data of the other's.
Therefore the \hr{raw} data's privacy are protected for any participant of knowledge graph owners.

As we incorporate the differential privacy mechanism in our \ph{PPAT network}, following PATE-GAN's proof, we can also estimate $\epsilon$ 
in Def. ~\ref{def:dp} for \ph{PPAT}.
We formulate the upper bound of $\hat{\epsilon}$ in \ph{PPAT} in Eq. (\ref{eq:epsilon}):

\noindent
\begin{equation}
\small
\label{eq:epsilon}
    \hat{\epsilon} = \min\limits_{l}\frac{\alpha(l)+\log(\frac{1}{\delta})}{l},
\end{equation}
where $\alpha(l)$ is the moments accountant for $l$-th moment. Its upper bound is derived by Theorem 2 and Theorem 3 in PATE mechanism \cite{nicolas_pate} as shown in Eq. (\ref{eq:alpha}):

\begin{equation}
\label{eq:alpha}
\small
    \alpha(l) = \alpha(l)+\min\left\{ 2\lambda^2 l(l+1), \log\left((1-q)\left( \frac{1-q}{1-e^{2\lambda}q} \right)^l + qe^{2\lambda l}\right) \right\},
\end{equation}
where $q$ is an intermediary value which is formulated in Eq.(\ref{eq:q}):
\begin{equation}
\label{eq:q}
\small
    q = \frac{2+\lambda |n_0 - n_1|}{4\exp(\lambda |n_0 - n_1|)},
\end{equation}
where $n_0$, $n_1$ denote the number of teachers' votes for 0 or 1 separately.
More details about updating $\alpha$ and $\hat{\epsilon}$ are shown in the Algo.~\ref{DE_algo}.
By choosing $\delta$ and $\lambda$, the $\hat{\epsilon}$ can be calculated.



\begin{algorithm}[t]
\footnotesize
\SetKwInOut{KIN}{Input}
\SetKwInOut{KOUT}{Output}
\caption{KGProcessor}\label{processor}
\KIN{The Host KG $g_j$}
\KOUT{Best embedding $E_b$}
\tcp{Start self iterative training to get the best knowledge graph embedding and the best score $S_b$}

$E_b \leftarrow g_j$.train();\

$S_b \leftarrow g_j$.test($E_b$);\ 

$g_j$.state  $\leftarrow$ Ready;\

\tcp{Start federated learning, and $Q$ is handshake signal Queue} 
$Q$.receive\_signal(); \tcp*[f]{Receiving signal as Host}\

\While{$g_j$.state = Ready}
{
    $U_p \leftarrow$ False; \tcp*[f]{Set the improvement identifier $U_p$ to false}\
    
    \While{$Q$\ is\ not\ empty}
    {
        $g_j$.state = Busy; \tcp*[f]{Start \ph{PPAT}, and set the state to busy}\
        
        $E_{t} \leftarrow$ ActiveHandshake($g_j$, $Q$.poll());\
        
        $g_j$.aggregation($E_{t}$);\
        
        \tcp{Start $g_j$ KGEmb-Update, after aggregating embedding}
        $T_e \leftarrow g_j$.train(); \
        
        $T_s\leftarrow g_j$.test($T_e$); \
        
        \tcp{Backtrack function is the embodiment of our backtrack mechanism} 
        $U_p,\ S_b,\ E_b \leftarrow g_j$.backtrack($T_s, S_b, T_e, E_b$);\
        
        $g_j$.state = Ready;\
    }
    \If{$U_p$\ =\ False\ \textbf{and}\ $Q$\ is\ empty}{
        \tcp{Both following states are configured by KG owner or activated by handshake signal}
        \If{$g_j$.state = Sleep}{
                $g_j$.sleep();\
        }
        \If{$g_j$.state = Ready}{
            \tcp{$g_j$ begins self iterative training}
           $T_e \leftarrow$ $g_j$.train(); \
           
           $T_s \leftarrow$ $g_j$.test($T_e$);\
           
           $U_p,\ S_b,\ E_b \leftarrow g_j$.backtrack($T_s, S_b, T_e, E_b$);\
        }
    }
    \If{$U_p$\ = True\ \textbf{and}\ $Q$\ is\ empty}{
    \tcp{Broadcast handshake signal to other KGs having aligned entities and relations}
          $g_j$.send\_handshake\_signal();\
    }
}
\Return Best embedding $E_b$ ;\
\end{algorithm}

\subsection{Federated Training}\label{sec:fed_train}
For multiple KGs, we construct \ph{PPAT networks} between any pair of $(g_i, g_j) \in \mathcal{KG}$ where $\mathcal{E}_j \cap  \mathcal{E}_i \neq \emptyset$ or $\mathcal{R}_j \cap  \mathcal{R}_i \neq \emptyset$, and produce $2\times\binom{\mathcal{N}}{2}$ \ph{PPAT networks} at most at the same time.
For any pair of $(g_i,g_j)$, at least one client and one host are required separately. 
Our asynchronous and decentralized setting allows individual KG owner to decide whether it should collaborate with other KGs.
The collaboration process can be described as a handshake protocol. 
Any $g_i$ has three states: \emph{Ready}, \emph{Busy}, and \emph{Sleep}.
\emph{Ready} state indicates $g_i$ having available computational resources and being active to pair up with other KGs. 
\emph{Busy} state indicates that $g_i$ does not have enough resources and will not respond to any collaboration request at the moment. 
Instead, collaborators will be put in a queue till $g_i$ finishes its work and is ready for collaborations.
\emph{Sleep} state indicates that though $g_i$ has the computational resources, it has not received any collaboration request yet. 
That is, If \emph{Ready} state cannot find a partner, it will switch to \emph{Sleep} state and wake up to \emph{Ready} state after a certain time period or being notified by a collaboration request.
A successful handshake process between $g_i$ and $g_j$ implies $state(g_i) \neq Busy, state(g_j) \neq Busy $, and at least one of them has \emph{Ready} state.
Algo. \ref{processor} describes how a KG refines its embeddings.
The whole handshake mechanism is explained in the Algo.~\ref{DE_algo}.

\begin{algorithm}[h]
{\footnotesize
\SetKwInOut{KIN}{Input}
\SetKwInOut{KOUT}{Output}
    \caption{ActiveHandshake}\label{DE_algo} %
    \KIN{The caller or Host KG $g_j$, Client KG $g_i$, Parameter $\delta$, Noise $\lambda$}
    \KOUT{Translated embedding $E_t$}
    \tcp{Initialization}
    $\mathbf{\ph{PPAT}}$.initialize($g_j, g_i$);\quad
    $\alpha=0$\;
    \Repeat{$\mathbf{\ph{PPAT}}$ \ training\ converged}{
        $AdvS \leftarrow g_i$.generate($X$);\tcp*[f]{Generate adversarial samples}\ 
        
        $g_j$.receive($g_i$.send($AdvS$));\tcp*[f]{Communication between processes}\
        
        \tcp{Train all teacher discriminators with partitioned $AdvS$}
        $Probs \leftarrow g_j$.teacher\_{D}s($Y, AdvS$);\
        
        $vote \leftarrow \text{PATE}_{\lambda}(Probs)$;\
        
        \tcp{Train student discriminator with noisily voted labels}
        $Proba \leftarrow g_j$.student\_{D}($vote, AdvS$);\
        
        $L_G, L_S \leftarrow g_j$.loss\_calculation($Proba$);\
        
        $ L_{T} \leftarrow g_j$.loss\_calculation($Probs$);\
        
        $grad$\_$G$, $grad$\_$T$\_$D$s, $grad$\_$S$\_
        $D \leftarrow$ backpropagation($L_G, L_{T}, L_S$);\
        
        $g_i$.receive($g_j$.send($grad\_G$)); 
        
        $g_j$.update\_parameters($\Theta_T, grad\_T\_Ds$);\
        
        $g_j$.update\_parameters($\Theta_S, grad\_S\_D$);\
        
        $g_i$.update\_parameters($\Theta_G, grad\_G$);\
        
        \tcp{Update moments}
        $q \leftarrow \frac{2+\lambda |n_0 - n_1|}{4\exp(\lambda |n_0 - n_1|)}$ ;\
        
        $\alpha(l) \leftarrow \alpha(l)+\min\left\{ 2\lambda^2 l(l+1), \log\left((1-q)\left( \frac{1-q}{1-e^{2\lambda}q} \right)^l + q e^{2\lambda l}\right) \right\}$;\

    }
    $\hat{\epsilon} \leftarrow \min\limits_{l}\frac{\alpha(l)+\log(\frac{1}{\delta})}{l}$;\
    
    $E_t \leftarrow g_j$.receive($g_i$.generate($X$), $g_i$.generate($N(X)$)); \
    
    \Return Translated embedding $E_t$; \
}

\end{algorithm}

\section{Experiments}\label{sec:expe}
In this section, we present extensive experiments to evaluate the performance of the proposed FKGE framework.
In Section \ref{Sec:Experimental Settings}, we will introduce basic experimental setup and metrics.
Then we will use FKGE to conduct both triple classification and link prediction experiments in Section \ref{sec:evaluation}.
Moreover, we will conduct ablation study to show the effectiveness of FKGE in Section \ref{sec:ablation}.
Finally, we will analyze time cost for our experiment in Section \ref{sec:time}.

\begin{table}[t]
	\footnotesize
	\caption{Statistics of the Knowledge Graphs.}\label{tab:dataset-table}\vspace{-0.1in}
	\resizebox{0.30 \textwidth}{!}{
       \begin{tabular}{@{}c|cc|c|c@{}}
			\toprule
			\multicolumn{2}{c|}{KGs} & \#Relation & \#Entity & \#Triple\\ \midrule
		    \multicolumn{2}{c|}{Dbpedia} & 14,085 & 49,1078 & 1,373,644 \\
		    \multicolumn{2}{c|}{Geonames} & 6 & 300,000 & 1,163,878 \\
		    \multicolumn{2}{c|}{Yago} & 37 & 286,389 & 1,824,322 \\
		    \multicolumn{2}{c|}{Geospecies} & 38 & 41,943 & 782,120 \\
		    \multicolumn{2}{c|}{Poképédia} & 28 & 238,008 & 548,883 \\
		    \multicolumn{2}{c|}{Sandrart} & 20 & 14,765 & 18,243 \\
		    \multicolumn{2}{c|}{Hellenic} & 4 &  11,145 & 33,296 \\
		    \multicolumn{2}{c|}{Lexvo} & 6 & 9,810 & 147,211 \\
		    \multicolumn{2}{c|}{Tharawat} & 12 & 4,693 & 31,130\\ 
		    \multicolumn{2}{c|}{Whisky} & 11 & 642 & 1,339 \\
		    \multicolumn{2}{c|}{World lift} & 10 & 357 & 1,192 \\\midrule
		    \multicolumn{2}{c|}{Summation} & 14,257 & 1,398,830 & 5,915,596\\
		    \bottomrule
	   \end{tabular}
	}\vspace{-0.1in}
\end{table}

\begin{table}[t]
	\footnotesize
	\caption{Statistics of Aligned Entities (AEs).}\label{tab:aligned-table}\vspace{-0.1in}
	\resizebox{0.38 \textwidth}{!}{
		\begin{tabular}{@{}c|cc|c|cc|c|c@{}}
			\toprule
			\multicolumn{2}{c|}{KGs} & \# AEs & \multicolumn{2}{|c|}{KGs} & \# AEs & \multicolumn{2}{|c}{KGs}\\\midrule
		    \multicolumn{2}{c|}{Geonames} & 118,939 & \multicolumn{2}{|c|}{Dbpedia} & 27 & \multicolumn{2}{|c}{Poképédia}\\ 
		    \multicolumn{2}{c|}{Yago} & 123,853 & \multicolumn{2}{|c|}{Dbpedia} & 133 & \multicolumn{2}{|c}{Geospecies}\\ 
		    \multicolumn{2}{c|}{Yago} & 53,553 & \multicolumn{2}{|c|}{Geonames} & 89 & \multicolumn{2}{|c}{Geospecies}\\ 
		    \multicolumn{2}{c|}{Sandrart} & 379 & \multicolumn{2}{|c|}{Dbpedia} & 41 & \multicolumn{2}{|c}{Hellenic}\\ 
		    \multicolumn{2}{c|}{Dbpedia} & 507 & \multicolumn{2}{|c|}{Lexvo} & 245 & \multicolumn{2}{|c}{Geonames}\\ 
		    \multicolumn{2}{c|}{Dbpedia} & 403 & \multicolumn{2}{|c|}{Tharawat 
		    } & 90 & \multicolumn{2}{|c}{Geonames}\\ 
		    \multicolumn{2}{c|}{Dbpedia} & 70 & \multicolumn{2}{|c|}{Whisky} & 39 & \multicolumn{2}{|c}{Geonames}\\ 
		    \multicolumn{2}{c|}{Dbpedia} & 25 & \multicolumn{2}{|c|}{World lift} & 18 & \multicolumn{2}{|c}{Yago}\\ 
		    \multicolumn{2}{c|}{Lexvo} & 77 & \multicolumn{2}{|c|}{Yago} & 266 & \multicolumn{2}{|c}{Tharawat 
		    }\\ 
		    \multicolumn{2}{c|}{Whisky} & 49 & \multicolumn{2}{|c|}{Yago} & - & \multicolumn{2}{|c}{-} \\\midrule
		\end{tabular}
	}
	\vspace{-0.1in}
\end{table}

\subsection{Experimental Settings}
\label{Sec:Experimental Settings}
\textbf{Dataset.} We select 11 KGs at different scales from the Linked Data community\footnote{https://lod-cloud.net/}.
For the KGE, OpenKE framework ~\cite{openke} is used so that FKGE is compatible with various KGE models.
For each KG, we count the numbers of relation, entity and triple, and divide the triples into train, valid, and test sets with ratio $90:5:5$ according to the default setting in the OpenKE.
Note that in order to reduce computational time of training and testing in KGE, we cut 
out some sparse entities and triples that are not relevant to aligned entities and triples from the original KGs.
A summary statistics of these KGs is shown in Tab. ~\ref{tab:dataset-table}. 
The Linked Data community provides \underline{a}ligned \underline{e}ntities (AE) between different KGs in RDF files, statistics as recorded in Tab. ~\ref{tab:aligned-table}. 

\begin{figure*}[h]
\centering
\setlength{\abovecaptionskip}{-0.0cm}
\subfigure[Baseline with TransE.]{\label{fig:baseline-transe}
\begin{minipage}[t]{0.32\linewidth}
\centering\vspace{-0.05in}
\includegraphics[width=6cm]{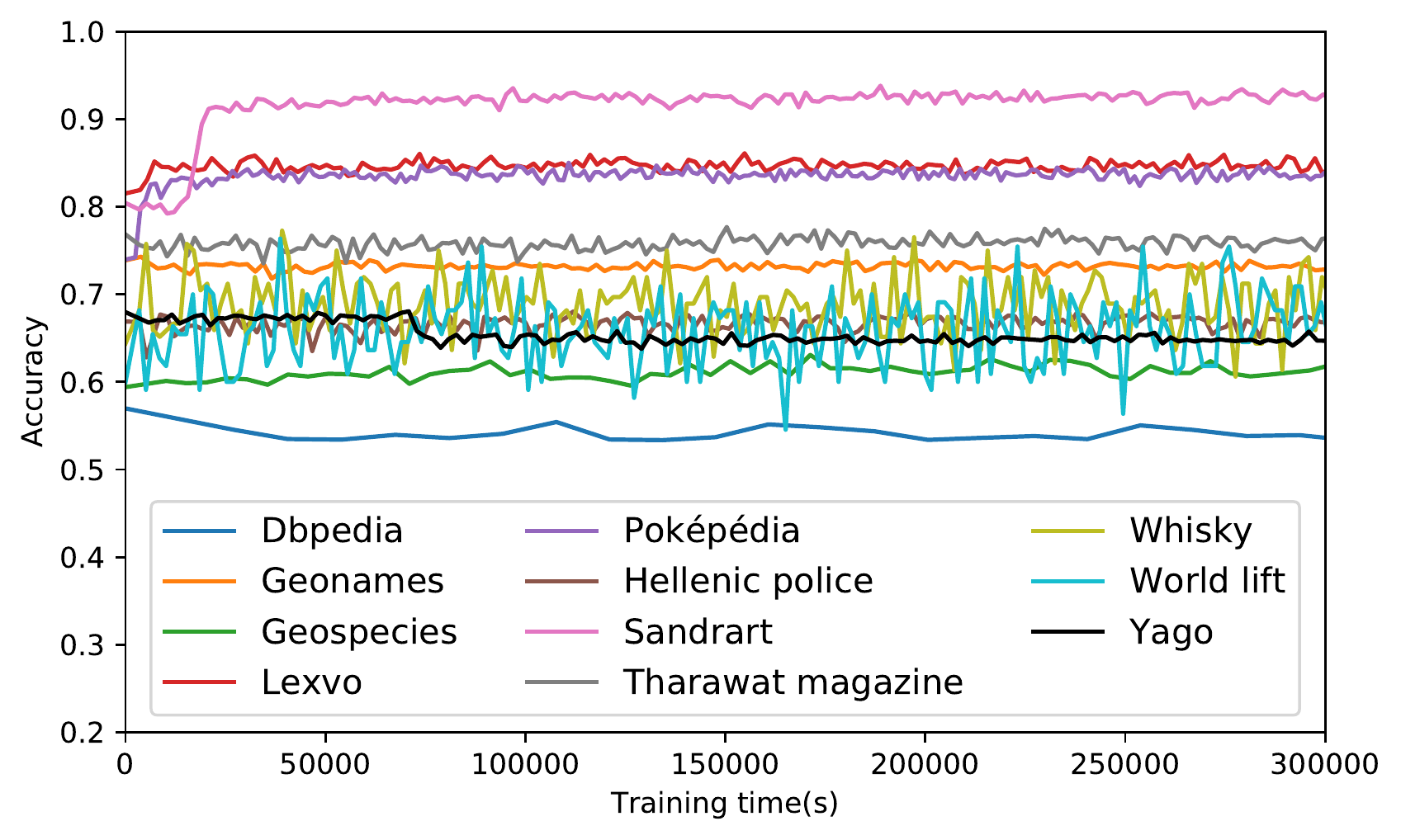}
\end{minipage}%
}
\subfigure[Baseline for unified KGE with TransE.]{\label{fig:unified}
\begin{minipage}[t]{0.32\linewidth}
\centering\vspace{-0.05in}
\includegraphics[width=6cm]{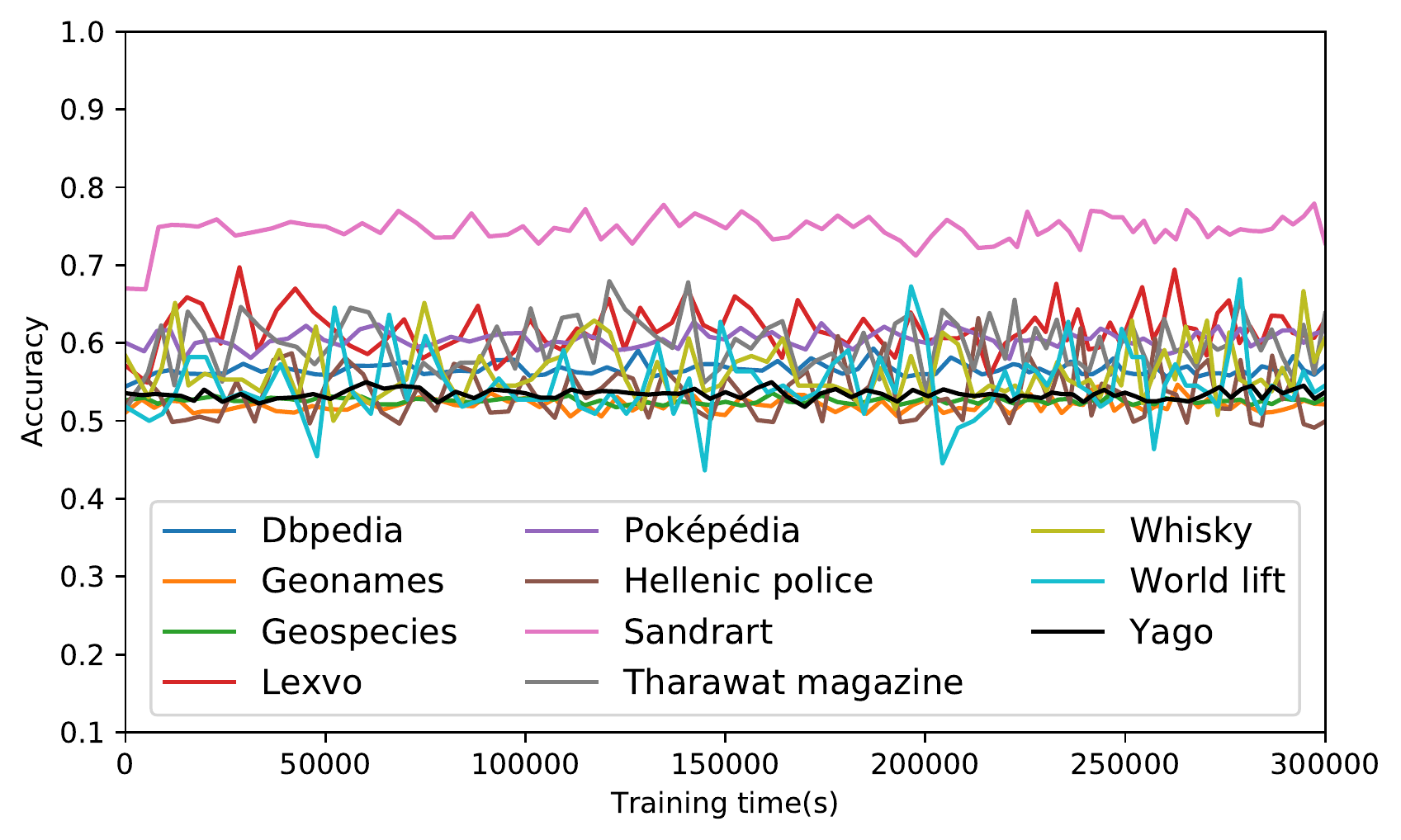}
\end{minipage}%
}
\subfigure[FKGE with TransE.]{\label{fig:fkge-transe}
\begin{minipage}[t]{0.32\linewidth}
\centering\vspace{-0.05in}
\includegraphics[width=6cm]{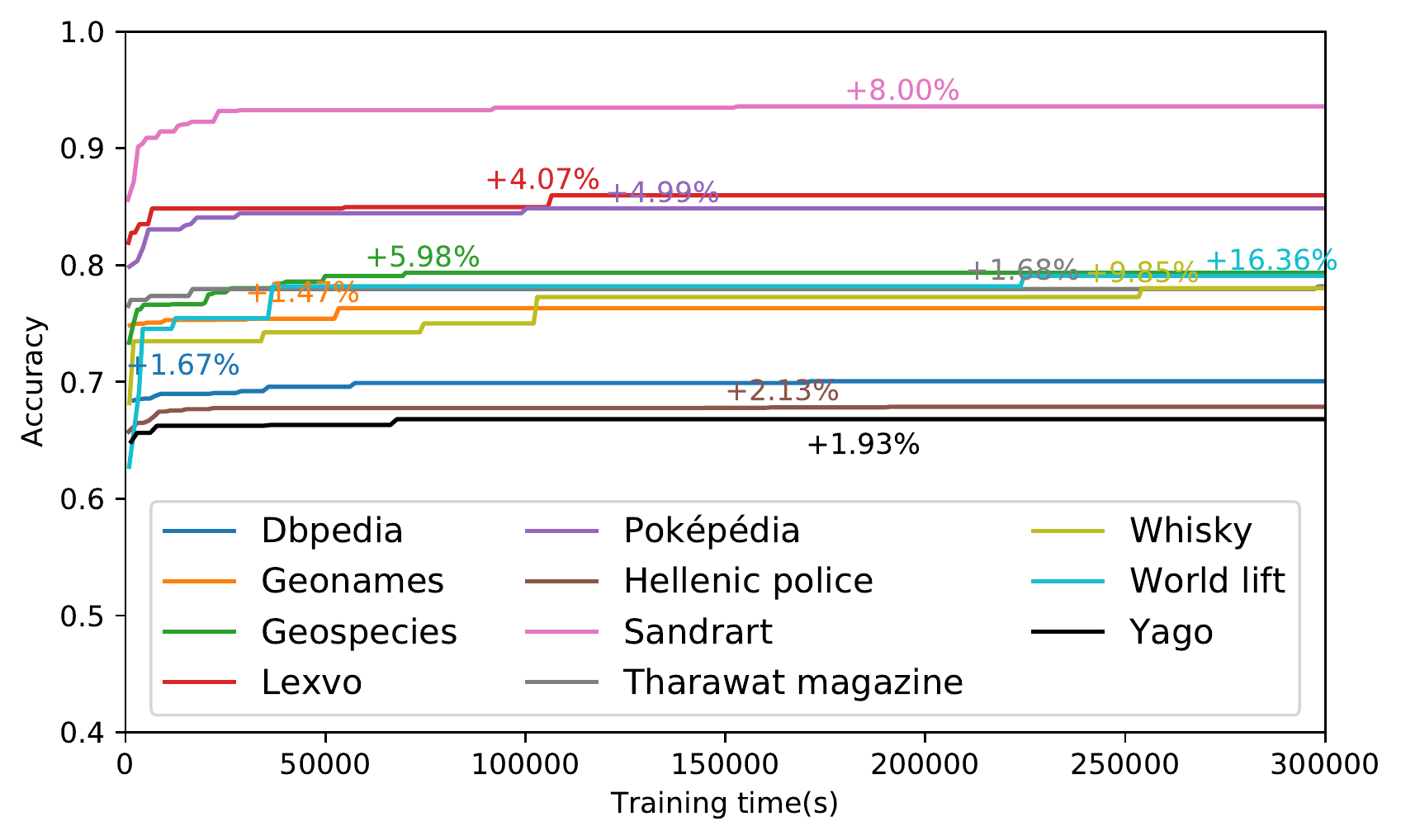}
\end{minipage}%
}
\subfigure[Baseline with different KGE models.]{\label{fig:fkge-baseline-more}
\begin{minipage}[t]{0.32\linewidth}
\centering\vspace{-0.05in}
\includegraphics[width=6cm]{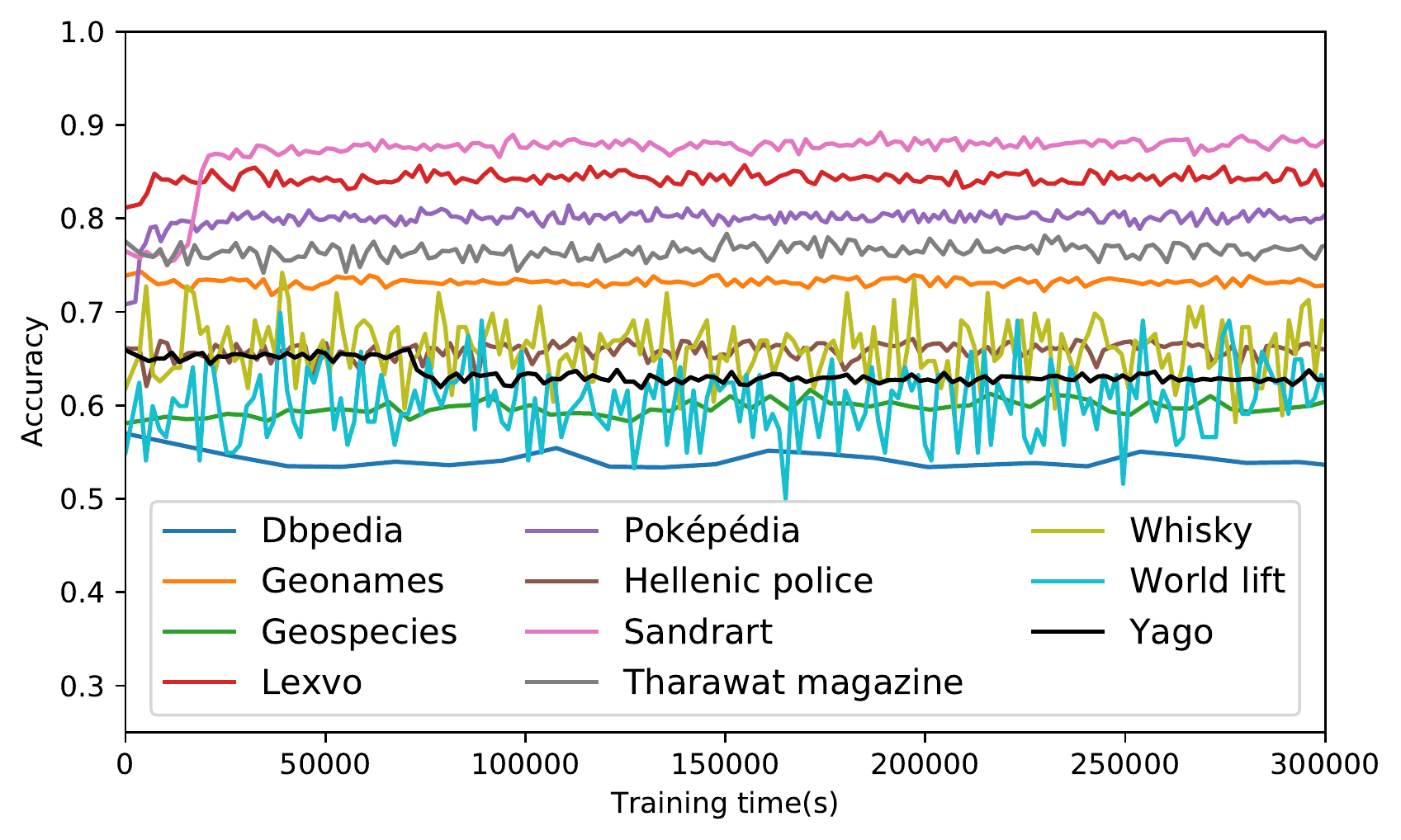}
\end{minipage}%
}
\subfigure[FKGE using different KGE models.]{\label{fig:fkge-trans-more}
\begin{minipage}[t]{0.32\linewidth}
\centering\vspace{-0.05in}
\includegraphics[width=6cm]{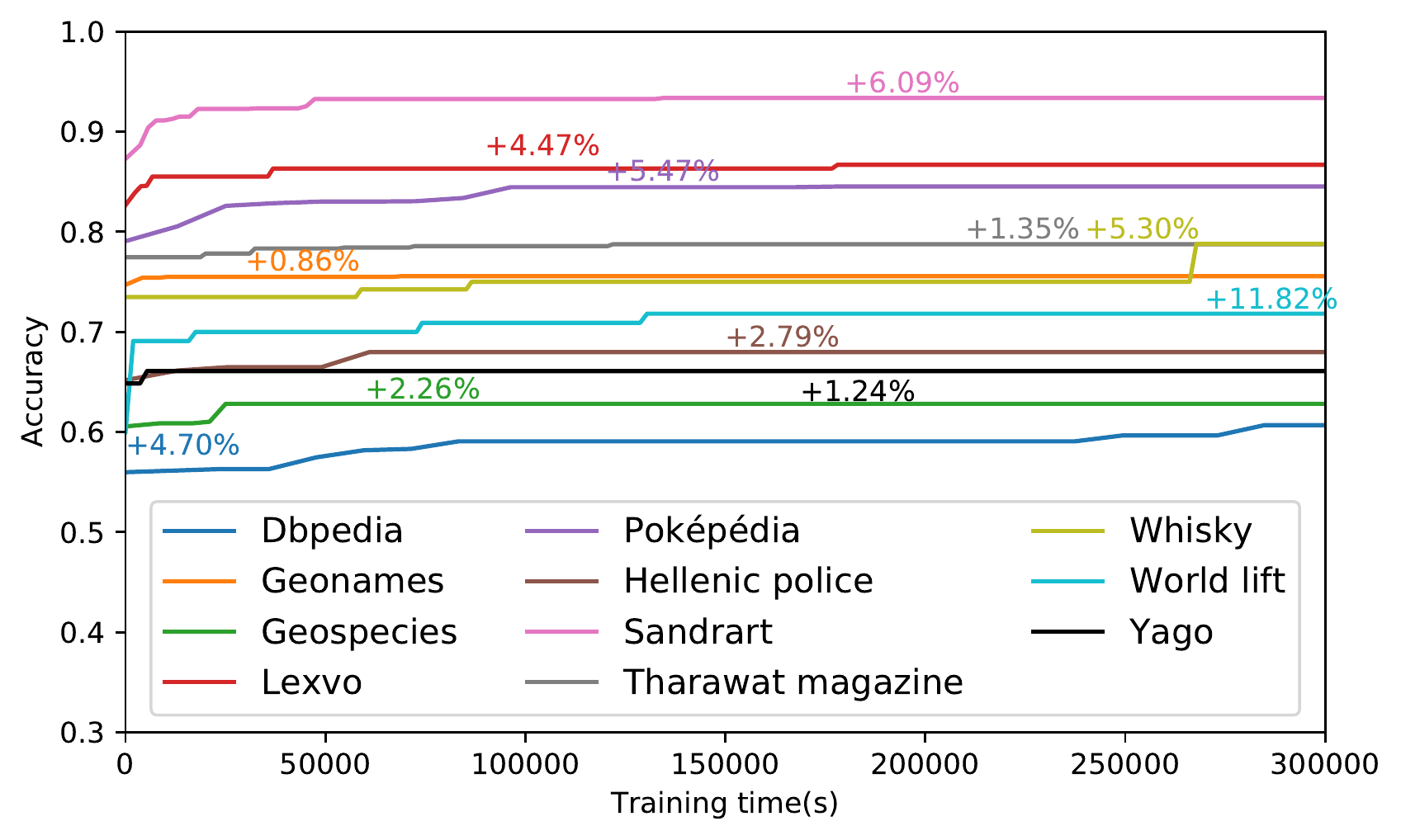}
\end{minipage}%
}
\subfigure[Experiment on manually divided \emph{Geonames}.]{\label{fig:fkge-entity-relation}
\begin{minipage}[t]{0.32\linewidth}
\centering\vspace{-0.05in}
\includegraphics[width=6cm]{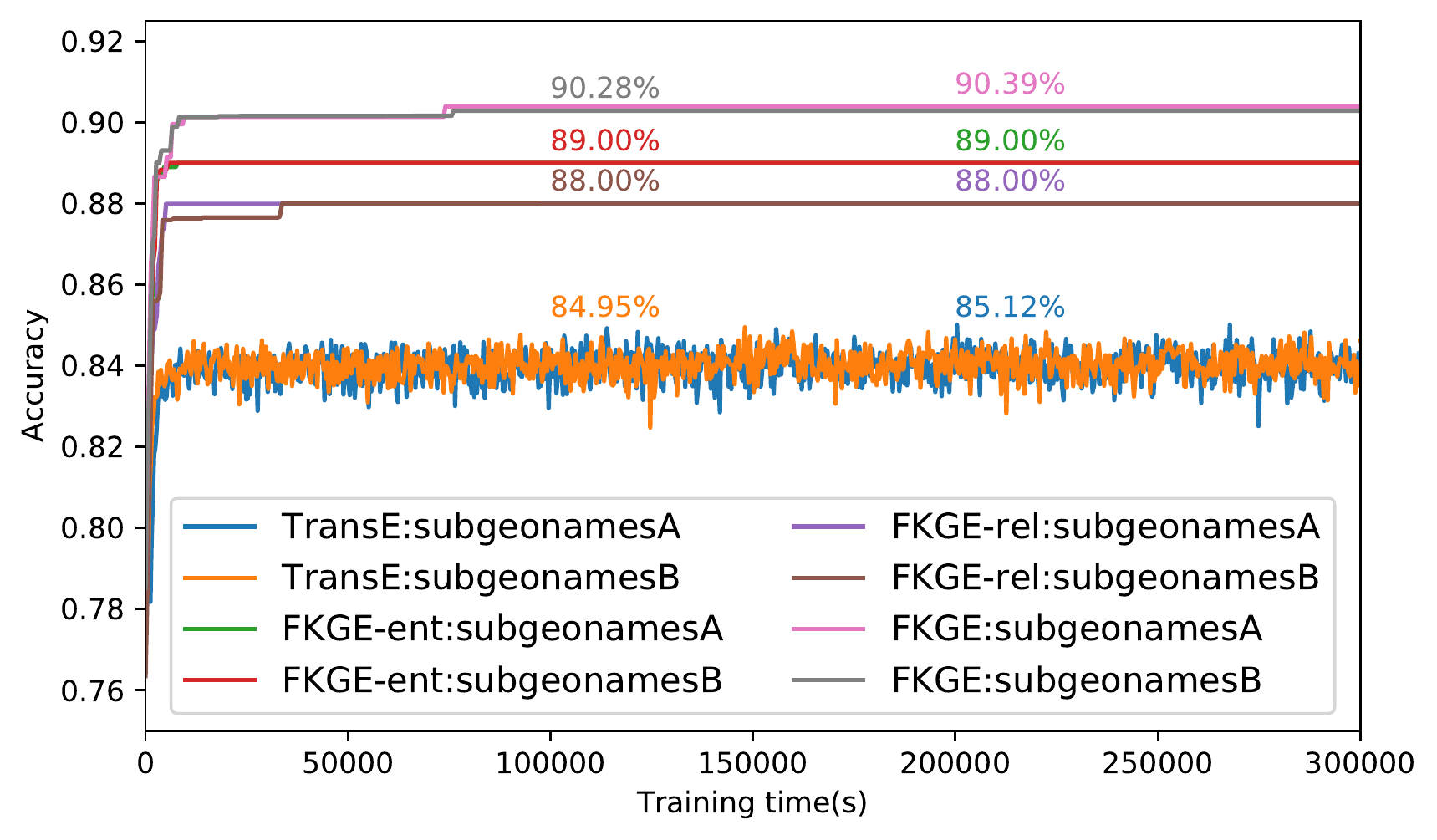}
\end{minipage}%
}
\centering
\caption{Performance of Triple Classification. }\label{fig:triple-classification}
\end{figure*}

\textbf{Hyperparameter Setting.}
To simulate real-world asynchronous training without data leakage, we set each KG to one process and implement all comparative experiments on 11 independent processes using the same type of GPU devices with the same configurations.
During a handshake process, we use pipeline communication between processes to transmit generated adversarial samples from client to host and gradients from host to client.
In consideration of computational time and testing results, we set default dimension $d$ of the embedding vector to $d = 100$, and step of testing to 1,000 epochs.
For other essential parameters of KGE, we set the learning rate to 0.5 and batch size to 100 following the default setting of the OpenKE ~\cite{openke}.
\hr{Negative samples are generated by corrupting either head or tail entities and ratios of negative samples and positive samples are 1:1.}
For the essential parameters of the \ph{PPAT network}, we set batch size, teacher number, learning rate, momentum to 32, 4, 0.02, 0.9.
We train for 1,000 epochs for each KG to get the initial best score and embeddings, and then activate the federal unit.

\textbf{Privacy Setting.}
We set $\lambda$ to 0.05 and $\delta$ to $10^{-5}$. 
For each pair of KGs in federated training, there is an \ph{PPAT network} with varied $\epsilon$ according to the input dataset $X$ since total the number of queries of paired embeddings may change according to number of aligned entities.
We estimate the largest upper bound $\hat{\epsilon}$ for all the $\epsilon$ and the bound is $\hat{\epsilon}=2.73$.
For each round of federated training, the max $\alpha(l)$ is 0.29 among all the ActiveHandshake in the Alg.~\ref{DE_algo}, putting $\ln{\frac{1}{\delta} }=11.5$ and $l=9$ together, we obtain upper bound $2.73$ for all $\epsilon$ following Eqs. (\ref{eq:epsilon} - \ref{eq:q}).

\textbf{Evaluation Metrics.}
The proposed FKGE framework is compatible with different KGE methods. 
We choose popular and simple translation-based models including TransE ~\cite{bordes2013translating}, TransH ~\cite{wang2014knowledge}, TransR ~\cite{lin2015learning}, and TransD ~\cite{ji2015knowledge} from the OpenKE ~\cite{openke} to evaluate the quality of the embeddings trained by different methods under two classical testing tasks of KGE: triple classification and link prediction.
For triple classification, we apply the \textbf{Accuracy} as evaluation metric.
For link prediction, we apply the widely used proportion of correct entities in top-1, 3 and 10 ranked entities (\textbf{Hit@1, 3 and 10}) and \textbf{Mean Rank} as evaluation metrics.

\subsection{Evaluation}\label{sec:evaluation}
Here, we demonstrate the advantages of the FKGE based on experimental results on triple classification and link prediction tasks.

\begin{table*}[t]
	\footnotesize
	\resizebox{0.95\textwidth}{!}{
		\begin{tabular}{@{}cc|ccc|ccc||ccc|cccc@{}}
   		    \toprule
			\multicolumn{2}{c|}{Methods} & \multicolumn{3}{|c|}{Independent-TransE}  & \multicolumn{3}{|c||}{FKGE} & \multicolumn{3}{|c|}{Random-Independent-KGE} & \multicolumn{3}{|c}{Multi-FKGE}\\\midrule
			\multicolumn{2}{c|}{Metric} & Hit@10 & Hit@3 & Hit@1 & Hit@10 & Hit@3 & Hit@1 & Hit@10 & Hit@3 & Hit@1 & Hit@10 & Hit@3 & Hit@1 \\ \midrule
		    \multicolumn{2}{c|}{Dbpedia} & 23.29 & 12.88 & 5.12  & \textbf{25.07} & \textbf{14.41} & \textbf{6.37}  & 5.46 &  2.51 & 1.10 & \textbf{6.67}  & \textbf{3.20} & \textbf{1.24} \\ 
		    \multicolumn{2}{c|}{Geonames}   & 8.82 & 3.69 & 1.93 &  \textbf{9.65} & \textbf{4.88} & \textbf{2.12} &  8.45 & 4.53 & 1.90 & \textbf{8.85} & \textbf{4.97} & \textbf{2.14} \\
		    \multicolumn{2}{c|}{Yago} & 2.05 & 0.76 & 0.25  & \textbf{2.59} & \textbf{0.88} & \textbf{0.29} & 2.03 & \textbf{0.75} & \textbf{0.24}  & \textbf{2.36} & \textbf{0.75} & \textbf{0.24} \\
		    \multicolumn{2}{c|}{Geospecies}   & 58.49 & 45.81 & 34.01  & \textbf{60.97}  & \textbf{46.95} & \textbf{35.03}  & 38.68 & 26.43 & 13.12 &  \textbf{40.92} & \textbf{28.04} & \textbf{14.38} \\
		    \multicolumn{2}{c|}{Poképédia}  & 38.14 & 29.04 & 19.31 &  \textbf{45.58} & \textbf{35.48} & \textbf{24.90}  & 34.22 &  25.13 & 16.43 &  \textbf{42.12} & \textbf{32.14} & \textbf{22.65} \\
		    \multicolumn{2}{c|}{Sandrart} & 87.39 & 83.16 & 67.18 &  \textbf{88.65} & \textbf{84.97} & \textbf{72.14} & 87.71 & 83.71 & 68.91 &  \textbf{87.99} & \textbf{84.22} & \textbf{69.69} \\
		    \multicolumn{2}{c|}{Hellenic}  & 32.18 & 21.87 & 18.96 &  \textbf{33.00} & \textbf{22.87} & \textbf{19.35} &  32.21 & 22.23 & 18.59 &  \textbf{32.82} & \textbf{22.59} & \textbf{19.44}\\
		    \multicolumn{2}{c|}{Lexvo} & 85.67 & 76.07 & 58.29 &  \textbf{87.35} & \textbf{77.74} & \textbf{62.90} &  84.21 & 75.82 & 58.09 &  \textbf{85.72} & \textbf{76.99} & \textbf{59.76} \\
		    \multicolumn{2}{c|}{Tharawat}  & 12.48 & 4.56 & 1.67 &  \textbf{13.45} & \textbf{5.26} & \textbf{2.19}  & 12.30 & 4.38 & 1.39 &  \textbf{12.55} & \textbf{5.21} & \textbf{1.77} \\
		    \multicolumn{2}{c|}{Whisky}  & 28.78 & 15.15 & 9.84 &  \textbf{35.60} & \textbf{18.93} & \textbf{10.60} &  28.78 & 18.93 & 12.87  & \textbf{30.12} & \textbf{19.45} & \textbf{12.92}  \\
		    \multicolumn{2}{c|}{World lift} & 45.76 & 24.57 & 7.62 &  \textbf{51.69} & \textbf{28.88} & \textbf{11.17} &  18.64 & 8.47 & 1.69 &  \textbf{18.85} & \textbf{9.32} & \textbf{2.54} \\
		    \toprule
		\end{tabular}
	}\caption{Evaluation results on link prediction (\%). We show the best results with boldface.
    }\label{tab:link-predict}
\vspace{-0.35in}
\end{table*}

\vspace{-0.05in}
\subsubsection{Triple Classification}\label{sec:triple}
We give the baseline accuracies of triple classification of the 11 KGs using TransE in Fig.~\ref{fig:baseline-transe}. 
The performances of baseline are unstable: accuracies of \emph{Yago} and \emph{Dbpedia} are even reduced.
To verify the performance of one unified structure of multiple KGs, we integrate 11 KGs into a united KG by merging aligned entities, and then test the performance of TransE on each KG independently.
The accuracy of triple classification of the unified KG is shown in Fig.~\ref{fig:unified}.
Compared with the independent KG embedding in Fig.~\ref{fig:baseline-transe}, the unified KG embedding even has generally decreased by 6.82\% - 17.63\%.
Hence, integrating the embeddings of multiple KGs into one unified vector space does not help to obtain effective representation learning of KGs.

We apply the FKGE framework to the 11 KGs with TransE.
The results of triple classification are shown in Fig. ~\ref{fig:fkge-transe}, where the marked improvements are compared with the results before training (at time 0).
After the same training time, it can be observed that the accuracy of each KG increases.
Specifically, compared with the baseline method in Fig. ~\ref{fig:baseline-transe}, the accuracy results (KG ordered as in Tab.~\ref{tab:dataset-table}) have been improved by 16.49\%, 2.98\%, 2.06\%, 17.85\%, 2.11\%, 0.60\%, 0.48\%, 0.77\%, 1.82\%, 12.88\% and 14.55\% on triple classification task, respectively.
The improvements in accuracy of the above 11 KGs benefit from the cross-knowledge embedding integration in the FKGE.
Moreover, the continuous and steady improvements also show the effectiveness of the backtrack mechanism in the FKGE.
Therefore, based on the FKGE framework and TransE, the 11 KGs have achieved consistent improvements in triple classification.

\begin{figure*}[t]
\centering
\setlength{\abovecaptionskip}{-0.0cm} 
\subfigure[The first part containing 4 knowledge graphs.]{
\begin{minipage}[t]{0.32\linewidth}
\centering\vspace{-0.05in}
\includegraphics[width=6cm]{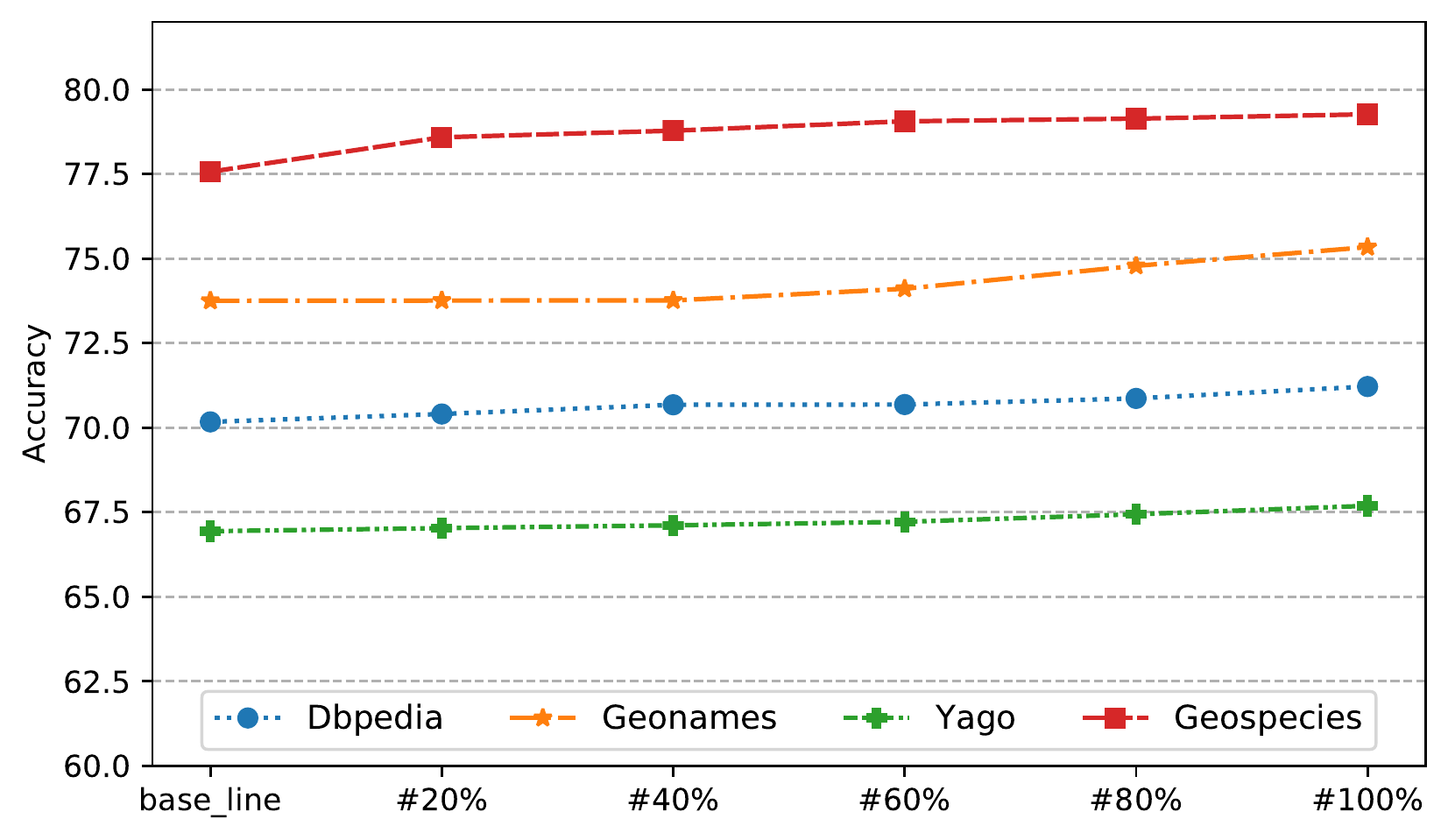}
\end{minipage}
}
\subfigure[The second part containing 4 knowledge graphs.]{
\begin{minipage}[t]{0.32\linewidth}
\centering\vspace{-0.05in}
\includegraphics[width=6cm]{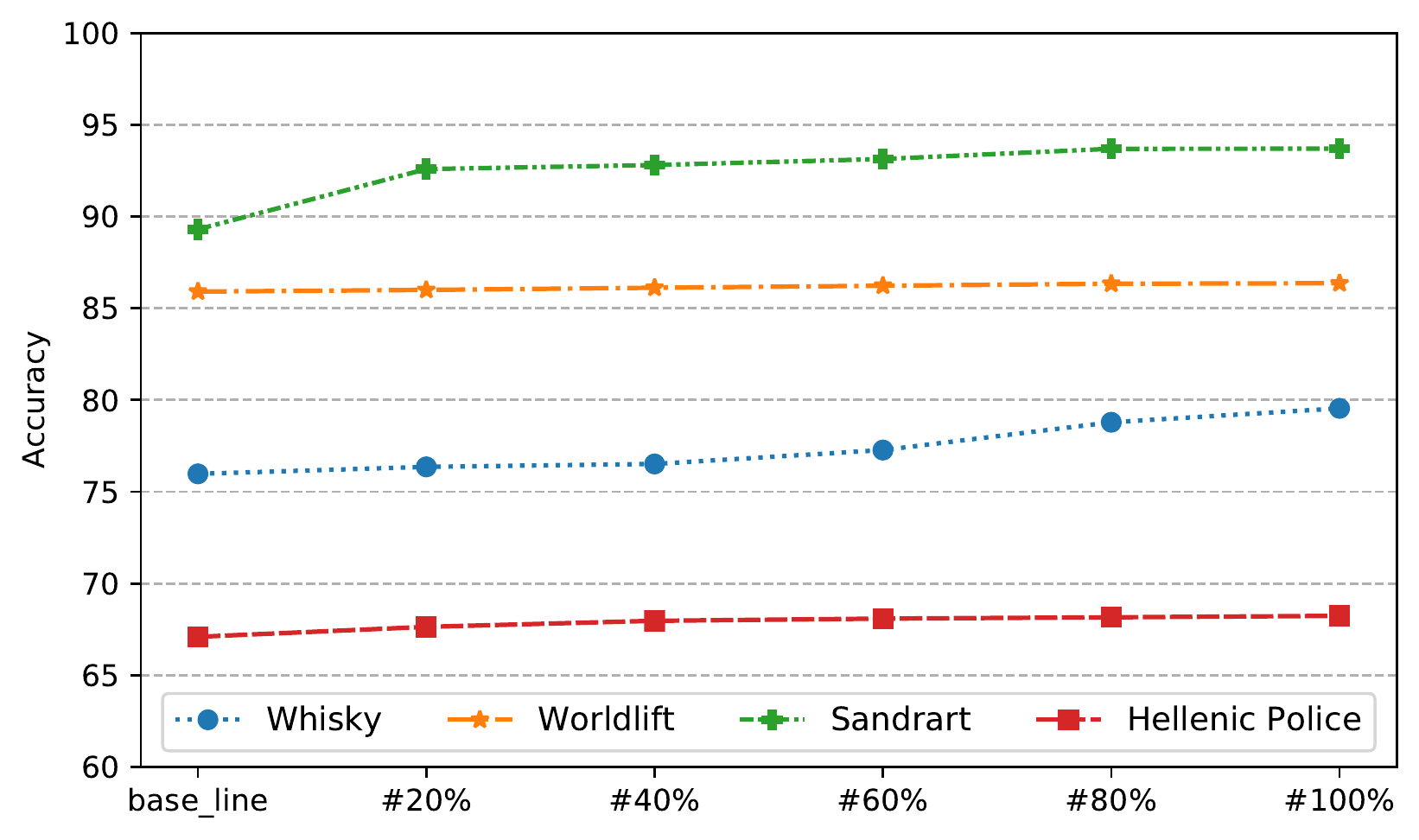}
\end{minipage}
}
\subfigure[The third part containing 3 knowledge graphs.]{
\begin{minipage}[t]{0.32\linewidth}
\centering\vspace{-0.05in}
\includegraphics[width=6cm]{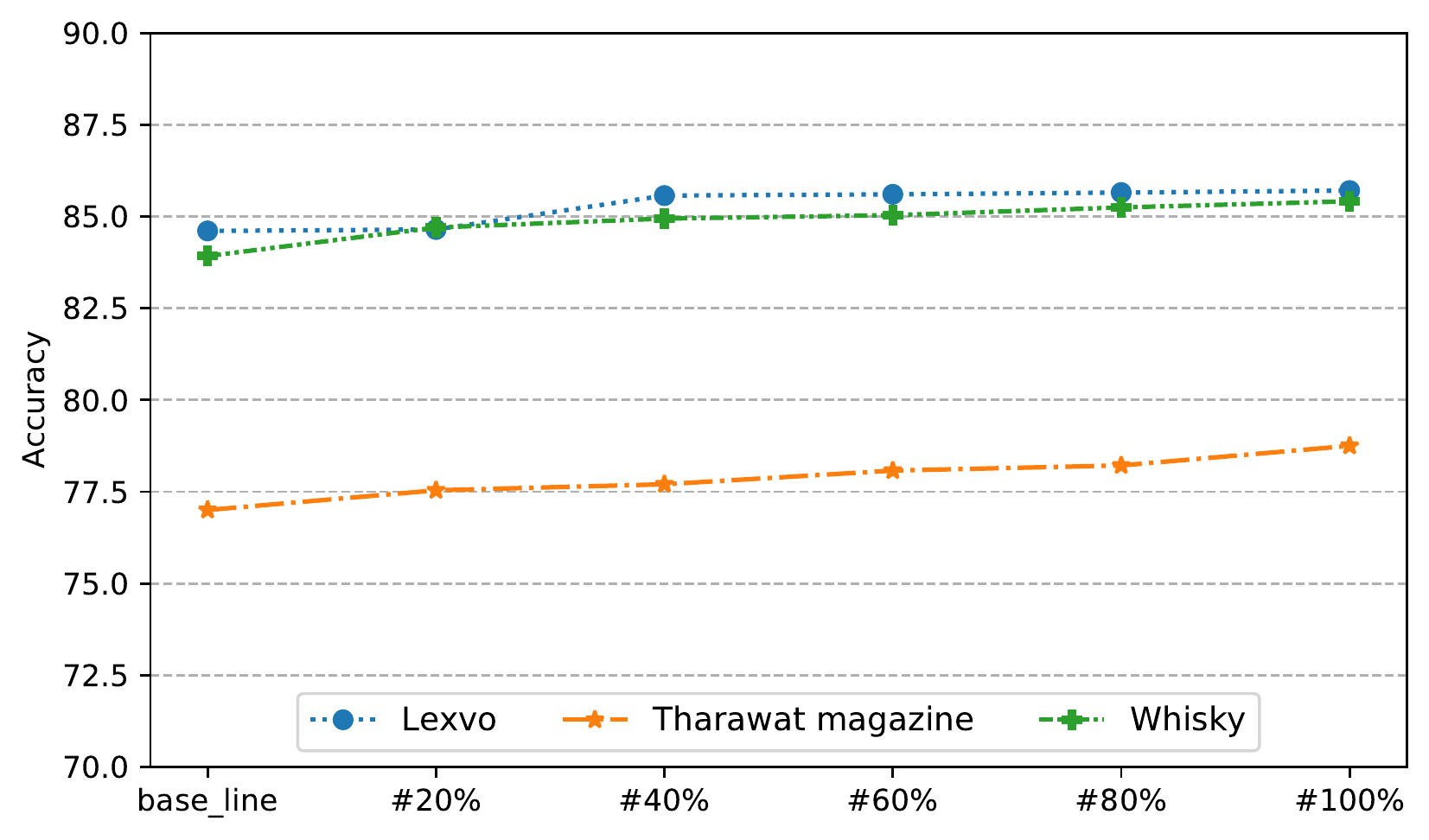}
\end{minipage}
}
\caption{Evaluation results on triple classification with different sampling rates.}\label{fig:ablation}
\vspace{-4mm}
\end{figure*}

\begin{table*}[t]
	\resizebox{1 \textwidth}{!}{
				\begin{tabular}{@{}c|c|cccc|cccc|cccc|cccc@{}}
			\toprule
			\multicolumn{2}{c|}{Sampling} & \multicolumn{4}{|c|}{\#20\%} & 	\multicolumn{4}{|c|}{\#40\%} & \multicolumn{4}{|c|}{\#60\%} & \multicolumn{4}{|c}{\#80\%}\\ \midrule
			\multicolumn{2}{c|}{Metric} & Mean Rank & Hit@10 & Hit@3 & Hit@1 & Mean Rank & Hit@10 & Hit@3 & Hit@1 & Mean Rank & Hit@10 & Hit@3 & Hit@1 & Mean Rank & Hit@10 & Hit@3 & Hit@1 \\ \midrule
		    \multicolumn{2}{c|}{Geonames} & 
		    63,080.34 & 8.93 & 3.86 & 1.93 & 
		    62,421.95 & 9.17 & 4.47 & 1.99 & 
		    61,276.67 & 9.35 & 4.49 & 2.04 & 
		    59,786.04 & 9.65 & 4.57 & 2.04  \\
		    \multicolumn{2}{c|}{Yago} & 
		    30,638.51 & 2.05 & 0.76 & 0.25 &
		    28,838.64 & 2.235 & 0.79 & 0.26 & 
		    26,563.40 & 2.35 & 0.81 & 0.26 & 
		    26,421.51 & 2.55 & 0.82 & 0.27 \\
		    \multicolumn{2}{c|}{Geospecies} &
		    415.47 & 59.31 & 45.97 & 34.81 & 
		    409.06 & 59.70 & 45.99 & 35.08 & 
		    400.73 & 60.17 & 46.38 & 35.12 & 
		    397.69 & 60.28 & 46.72 & 35.30 \\
		    \multicolumn{2}{c|}{Poképédia} & 
		    7,463.42 & 38.37 & 30.01 & 20.01 & 
		    7,450.85 & 43.25 & 30.50 & 21.61 & 
		    7,445.54 & 44.09 & 33.08 & 23.39 & 
		    7,434.55 & 45.49 & 33.43 & 24.19 \\
		    \multicolumn{2}{c|}{Sandrart} & 
		    56.23 & 87.61 & 83.38 & 67.96 & 
		    52.39 & 88.38 & 83.43 & 69.51 & 
		    51.71 & 88.65 & 84.34 & 70.38 & 
		    47.39 & 88.98 & 84.37 & 72.10 \\
		    \multicolumn{2}{c|}{Hellenic Police} & 
		    1,886.24 & 32.29 & 22.06 & 19.03 & 
		    1,853.55 & 32.47 & 22.41 & 19.08 & 
		    1,823.84 & 32.63 & 22.78 & 19.11 & 
		    1,751.37 & 32.98 & 22.86 & 19.27 \\
		    \multicolumn{2}{c|}{Lexvo} &
		    43.64 & 86.01 & 76.59 & 58.43 & 
		    44.96 & 86.14 & 76.86 & 59.68 & 
		    43.43 & 86.70 & 77.11 & 60.47 & 
		    41.57 & 86.77 & 77.62 & 61.30 \\
		    \multicolumn{2}{c|}{Tharawat magazine} & 
		    232.63 & 12.50 & 4.58 & 1.76 & 
		    232.62 & 12.92 & 4.81 & 1.96 & 
		    228.65 & 13.13 & 5.09 & 1.99 & 
		    226.67 & 13.19 & 5.15 & 2.10 \\
		    \multicolumn{2}{c|}{Whisky} & 
		    49.47 & 29.39 & 15.98 & 9.91 & 
		    49.85 & 31.14 & 16.24 & 10.15 & 
		    44.47 & 31.85 & 17.06 & 10.25 & 
		    43.12 & 33.35 & 17.64 & 10.58 \\
		    \multicolumn{2}{c|}{World lift} &
		    22.98 & 46.80 & 25.47 & 7.85 & 
		    22.82 & 49.71 & 25.60 & 7.99 & 
		    22.74 & 50.06 & 28.62 & 8.02 & 
		    22.11 & 50.15 & 28.87 & 10.30 \\
		    \toprule
		\end{tabular}
	}\caption{Evaluation results on link prediction tasks in Filter with different sampling ratio(\%) of aligned entities.}\label{tab:link-scale-filter}
\vspace{-5mm}
\end{table*}

Not only TransE benefits from the \ph{PPAT network}, but also other mainstream KGE models can be improved through the FKGE framework.
We also randomly select KGE method from 4 popular translation family models, including TransR, TransE, TransD and TransH, for each KG.
We give the results  of the baseline using different translation family models in Fig.~\ref{fig:fkge-baseline-more}.
After 300,000 seconds of training, the accuracy of the FKGE in the triple classification for the 11 KGs is shown in Fig.~\ref{fig:fkge-trans-more}, where the marked improvements are also compared with the results before training (at time 0).
Specifically, compared with their respective base methods, the 11 KGs have been improved by 7.08\% (TransR), 2.23\% (TransD), 1.33\% (TransE), 1.32\% (TransR), 1.77\% (TransE), 0.38\% (TransD), 0.57\% (TransD), 1.92\% (TransD), 2.42\% (TransD), 13.64\% (TransH), and 7.27\% (TransR) on the triple classification, respectively. 
It confirms that the FKGE framework has the advantage of being compatible with different KGE methods.

\vspace{-0.1in}
\subsubsection{Link Prediction}\label{sec:link prediction}
We compare the performance of link prediction with type constraint following the OpenKE in multiple scenarios.
As shown in Tab. ~\ref{tab:link-predict}, we present the evaluation results of Hit@1, 3, and 10 in the Filter setting. 
The Filter means that those corrupted triples in test set and validation set are removed for the link prediction.
The Independent-TransE means using the traditional TransE based KGE individually.
\hr{The FKGE means using TransE and federated training to improve every KG collaboratively.
Besides TransE, we also use other KGE methods for link prediction.}
The Random-Independent-KGE means that each KG is applied with a KGE randomly from translation-based models and is training independently.
The Multi-FKGE means that each KG randomly chooses a KGE from the translation-based models and also employs the FKGE for further training.
Here, we keep the same base KGE methods selected in Fig. ~\ref{fig:fkge-trans-more}.
Compared with the baseline methods, the TransE based FKGE has increased at most by 7.44\%, 6.44\%, and 5.59\%  in terms of Hit@10, 3, and 1, and Multi-FKGE has increased at most by 7.90\%, 7.01\%, and 5.87\%  in terms of Hit@10, 3, and 1.
For example, benefited from the TransE based FKGE, the \emph{World lift} gains 5.93\%, 4.31\%, and 3.55\% improvements in terms of Hit@10, 3, and 1 under the Filter.
The above experiments under link prediction also demonstrate the effectiveness and adaptability of the proposed FKGE framework.

\vspace{-0.05in}
\subsection{Ablation Study}\label{sec:ablation}
\textbf{Effectiveness of aligned entities and relations}.
Here we first consider whether the inclusion of aligned entities and relations is beneficial to FKGE's performance.
Because the existing KGs provide no aligned relations, we manually divide \emph{Geonames} into two subsets named \emph{SubgeonamesA} and \emph{SubgeonamesB} of the same size to verify the contribution of aligned entities and relations.
We treat relations similarly as entities and simply put them together for model training.
Note that the size of divided KGs shrinks on both entity and relation, and are tested on different sets of triples. 
Therefore the accuracies may differ from baselines in Fig.~\ref{fig:baseline-transe}.
As shown in Fig. ~\ref{fig:fkge-entity-relation}, the blue and orange lines represent the accuracy of the two subsets with TransE only, namely \emph{TransE: subgeonamesA} and \emph{TransE: subgeonamesB}, in the triple classification; 
the green and red lines represent the accuracy by using only the aligned entities based on FKGE with TransE, namely \emph{FKGE-ent: subgeonamesA} and \emph{FKGE-ent: subgeonamesB}; 
the purple and brown lines represent the accuracy by using only the aligned relations based on FKGE with TransE, namely \emph{FKGE-rel: subgeonamesA} and \emph{FKGE-rel: subgeonamesB}; 
the pink and grey lines represent the accuracy by using both aligned entities and relations based on FKGE with TransE, namely \emph{FKGE: subgeonamesA} and \emph{FKGE: subgeonamesB}, respectively.
Obviously, compared with the \emph{TransE: subgeonamesA} and \emph{TransE: subgeonamesB}, \emph{FKGE: subgeonamesA} and \emph{FKGE: subgeonamesB} achieve significant improvements of 5.33\% and 5.27\% in triple classification, respectively.
Besides that, both \emph{FKGE-ent: subgeonamesA} and \emph{FKGE-ent: subgeonamesB} achieve 1.00\% improvement over the relation alignment based \emph{FKGE-rel: subgeonamesA} and \emph{FKGE-rel: subgeonamesB}.
In general, for the proposed FKGE framework, aligned entities and relations both contribute to the improvement of knowledge graph representations, and the refined embeddings show that \ph{PPAT network} is also effective.

\textbf{Size of aligned entities}.
In order to measure the impact of the scale of aligned entities involved in the FKGE, we compare the performance of different numbers of the aligned entities and different numbers of KGs under triple classification.
We randomly sample 20\%, 40\%, 60\% and 80\% of the aligned entities for fusing them through the \ph{PPAT network}, respectively.
The results on triple classification with different sampling rates are shown in Fig.~\ref{fig:ablation}.
Intuitively, the more entities' embeddings that can be fused, the more obvious the improvements are.
For example, by sampling 20\%, 40\%, 60\% and 80\% and 100\% sampling rates, the \emph{World lift} gains 3.52\%, 4.05\%, 5.19\%, 6.44\% and 7.16\% improvements on the triple classification task, respectively.
Besides triple classification, Tab.~\ref{tab:link-scale-filter} presents the results of link prediction by using TransE based FKGE, in terms of Mean Rank, Hit@10, Hit@3 and Hit@1, with different sampling ratio in the \ph{PPAT network}.
In the Filter metric, by using the FKGE of 20\%, 40\%, 60\%, 80\% and 100\% (Presented in Tab.~\ref{tab:link-predict}) sampling rates, the \emph{Poképédia} gains up to 0.23\%, 5.11\%, 5.95\%, 7.35\% and 7.44\% improvements in terms of Hit@10 on the link prediction task, respectively.
In addition to the metric of Hit@10, most of other KGs have also achieved improvements in Mean Rank, Hit@3 and Hit@1.
The above experimental results have once again proved the scalability and effectiveness of the proposed FKGE framework.

\vspace{-0.05in}
\subsection{Time Consumption}\label{sec:time}
\begin{figure}[t]
\center
\includegraphics[width=0.4\textwidth,height=0.2\textwidth]{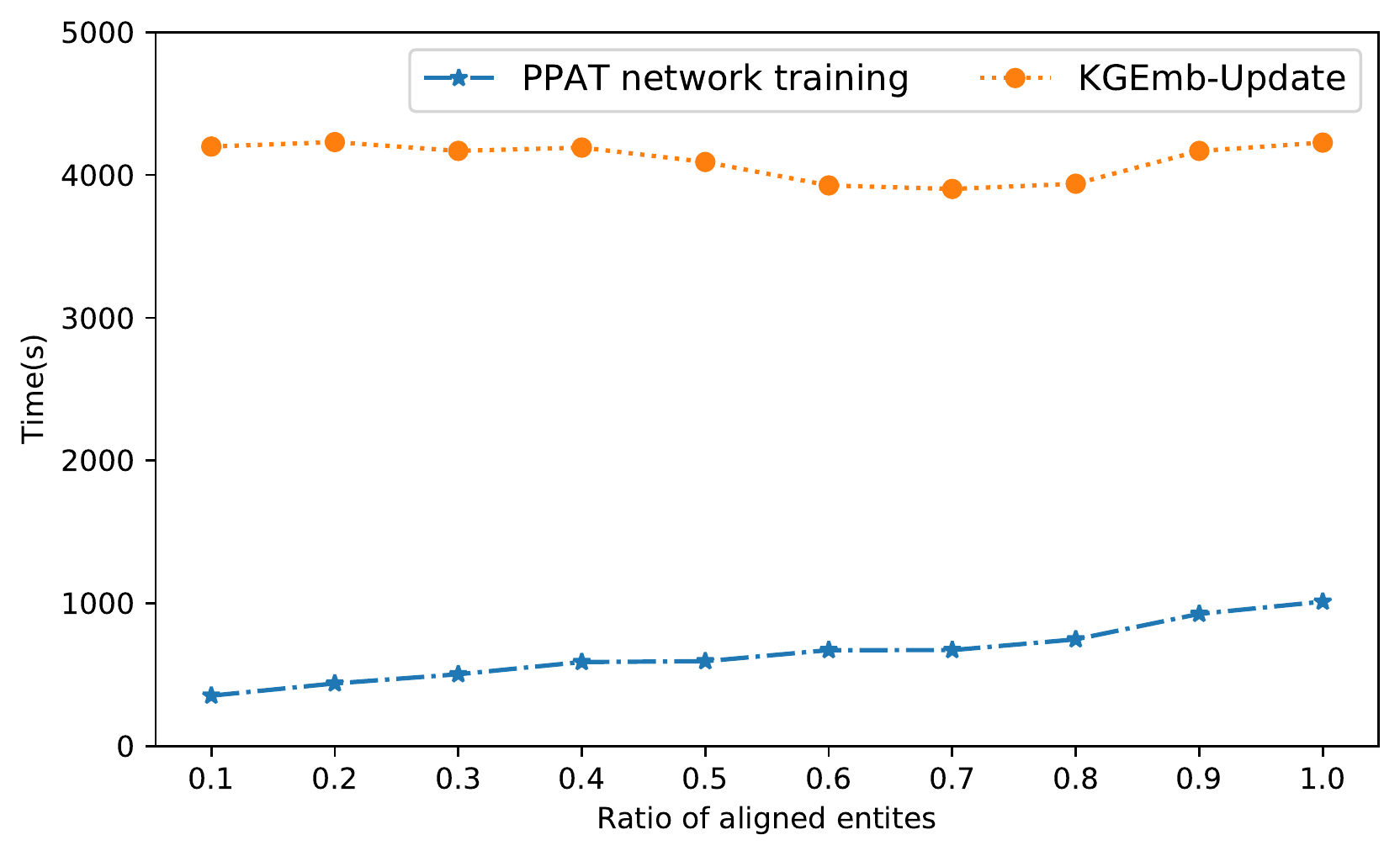}
\vspace{-0.15in}
\caption{Time cost experiment on \emph{Geonames} (in seconds).}\label{fig:dbpe_geoname_time}
\vspace{-0.15in}
\end{figure}

To analyze the time cost for individual KG and demonstrate scalability of FKGE.
During federation, we evaluate the time cost for different ratios of aligned entities between \emph{Geonames} and \emph{Dbpedia}.
We keep the exact experimental setup as Section \ref{Sec:Experimental Settings} except that only \emph{Geonames} and \emph{Dbpedia} are used for the experiment.
It's easy to verify that KGEmb-Update and \ph{PPAT} training pairwisely constitute major time costs for individual KGs.
Both of the KGs share similar results, and we show time consumption of \emph{Geonames} in Fig. \ref{fig:dbpe_geoname_time}.
\hr{For each ratio, we run the experiments for 10 times and average the corresponding time consumption.}
It can be observed that KGEmb-Update usually costs much more time than \ph{PPAT network} and its cost remains around 4,000s as number of aligned entities increases.
For \ph{PPAT} training, the cost increases roughly linearly from 350s to 1,000s as number of aligned entities increases, which constitutes around 8\% $\sim$ 20\% \hr{among} overall training time.
In general, for each KG in FKGE, overall time cost is acceptable compared with gained improvements. 
The linear time cost of the \ph{PPAT network} indicates FKGE is scalable for aligned entities. 
Moreover, for transmission between client and host, the client sends translated embeddings of size (batch size, $d$) and host transmits gradients for client of size $(d, d)$.
Using our experimental settings batch size $ = 32$, $d=100$ and 64 bit for double precision, total communication cost for a batch training of the  \ph{PPAT network} is at most 0.845 Mb. 
Therefore, it is practical for FKGE to train \ph{PPAT networks} online like a peer-to-peer network.

\section{Conclusion}\label{sec:conclu}
In this paper, we present a well-performed and decentralized federated knowledge graph embedding framework.
First, we exploit a privacy-preserving adversarial model between pairs of knowledge graphs to merge identical entities or relations of different domains, and guarantee no raw data leakage.
Then, we implement an asynchronous and pairwise federated representation learning framework on knowledge graphs.
We conduct extensive experiments to evaluate the FKGE on 11 knowledge graphs, demonstrating significant and consistent improvements in model quality through performance on triple classification and link prediction tasks.
The ablation study also indicates that merging both aligned entities and relations are beneficial for overall improvement.
Besides, the time consumption demonstrates the scalability of FKGE.
In the future, we plan to integrate more advanced knowledge graph representation learning models and extend our federated learning framework to other complex knowledge data exchanging scenarios.

\section*{Acknowledgment}
The authors of this paper were supported by the NSFC through grants U20B2053 and 62002007, the S\&T Program of Hebei through grant 20310101D, the Fundamental Research Funds for the Central Universities, the National Key Research and Development Program of China (208AAA0101100), the RIF (R6020-19 and R6021-20) and the GRF (16211520) from RGC of Hong Kong, the MHKJFS (MHP/001/19) from ITC of Hong Kong.
We thank Linfeng Du, Qi Teng, and Lichao Sun for useful comments and discussions, and also acknowledge the support from the HKUST-WeBank Joint Lab at HKUST.
For any correspondence, please refer to Hao Peng.

\bibliographystyle{ACM-Reference-Format}
\bibliography{acmart}

\end{document}